\documentclass[lettersize,journal]{IEEEtran}
\usepackage{amsmath,amsfonts}
\usepackage{algorithm}
\usepackage{algpseudocode} 
\usepackage{array}
\usepackage{url}
\usepackage{verbatim}
\usepackage{graphicx}
\usepackage{framed,multirow}
\usepackage{subfigure}
\usepackage[colorlinks=true, linkcolor = blue, citecolor = cyan]{hyperref}
\usepackage{textcomp}
\usepackage{cite}
\usepackage{multirow}
\usepackage[abbreviations]{glossaries-extra}

\newcommand{\vg}{\boldsymbol{g}}

\newcommand{\vm}{\boldsymbol{m}}

\newcommand{\vv}{\boldsymbol{v}}
\newcommand{\vx}{\boldsymbol{x}}
\newcommand{\vy}{\boldsymbol{y}}

\newcommand{\vtheta}{\boldsymbol{\theta}}
\newcommand{\muindex}{^{(\mu)}}
\newcommand{\muindexpre}{^{(\mu-1)}}
\begin{document}

\title{A Survey of Incremental Transfer Learning: Combining Peer-to-Peer Federated Learning and Domain Incremental Learning for Multicenter Collaboration}

\author{Yixing Huang, Christoph Bert, Ahmed Gomaa, Rainer Fietkau, Andreas Maier, \IEEEmembership{Senior Member, IEEE}, \\Florian Putz
\thanks{Y. Huang, C. Bert, A. Gomaa, R. Fietkau, and F. Putz are with Department of Radiation Oncology, University Hospital Erlangen, Friedrich-Alexander-Universit\"at Erlangen-N\"urnberg, 91054 Erlangen, Germany. They are also with Comprehensive Cancer Center Erlangen-EMN (CCC ER-EMN), 91054 Erlangen, Germany. Y. Huang, A. Gomaa, R. Fietkau, and F. Putz are also with Bavarian Cancer Research Center (BZKF), Erlangen, Germany.}
\thanks{A. Maier is with Pattern Recognition Lab, Friedrich-Alexander-Universit\"at Erlangen-N\"urnberg, 91058 Erlangen, Germany.}
}



\maketitle

\begin{abstract}
Due to data privacy constraints, data sharing among multiple clinical centers is restricted, which impedes the development of high performance deep learning models from multicenter collaboration. Naive weight transfer methods share intermediate model weights without raw data and hence can bypass data privacy restrictions. However, performance drops are typically observed when the model is transferred from one center to the next because of the forgetting problem. Incremental transfer learning, which combines peer-to-peer federated learning and domain incremental learning, can overcome the data privacy issue and meanwhile preserve model performance by using continual learning techniques. In this work, a conventional domain/task incremental learning framework is adapted for incremental transfer learning. A comprehensive survey on the efficacy of different regularization-based continual learning methods for multicenter collaboration is performed. The influences of data heterogeneity, classifier head setting, network optimizer, model initialization, center order, and weight transfer type have been investigated thoroughly. Our framework is publicly accessible to the research community for further development.
\end{abstract}


\section{Introduction}
\label{sect:Intro}
\IEEEPARstart{D}{eep} learning has played important roles in various medical applications nowadays, including image registration \cite{fu2020deep}, reconstruction \cite{huang2021data}, and segmentation \cite{weissmann2023deep,huang2021deep} as well as disease diagnosis and treatment planning \cite{avanzo2020radiomics}.
The performance of deep learning algorithms relies highly on the amount and quality of training data. Many research centers have the required technological and human resources as well as computation power. However, the limited access to data becomes an obstacle for them to develop deep learning algorithms independently. Therefore, collaboration among different centers (including research centers and hospitals) is always important. Due to data privacy and data management regulations, e.g., the EU medical device regulation \cite{beckers2021eu} and the EU General Data Protection Regulation for health data, data sharing among multiple centers is restricted, which impedes the development of high performance deep learning models from multicenter collaboration.

\begin{figure}
\centering
\begin{minipage}[b]{0.58\linewidth}
\subfigure[Center-to-peer]{
\includegraphics[width=\linewidth]{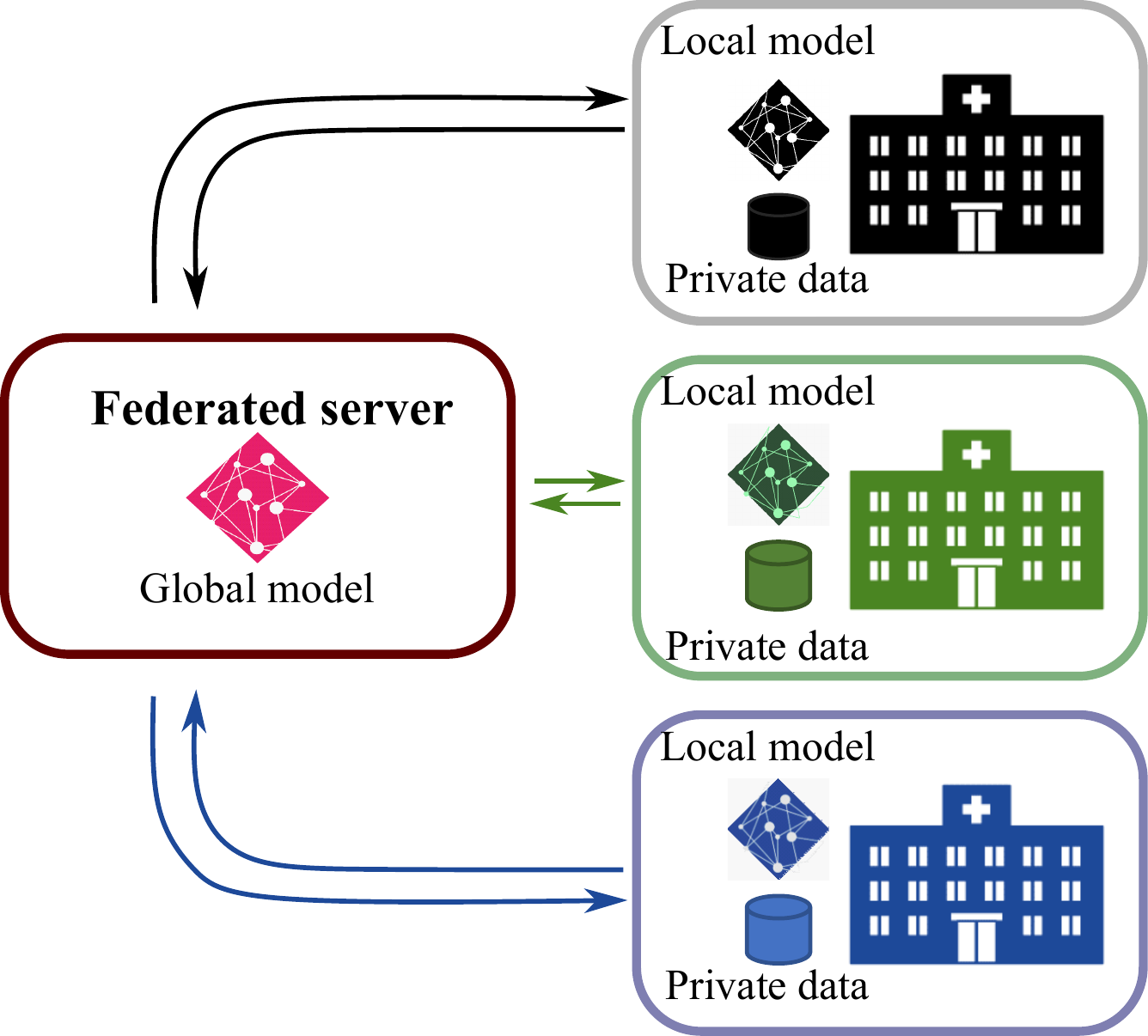}
\label{subfig:Center-to-client}
}
\end{minipage}
\begin{minipage}[b]{0.05\linewidth}
\ 
\end{minipage}
\begin{minipage}[b]{0.31\linewidth}
\subfigure[Peer-to-peer]{
\includegraphics[width=\linewidth]{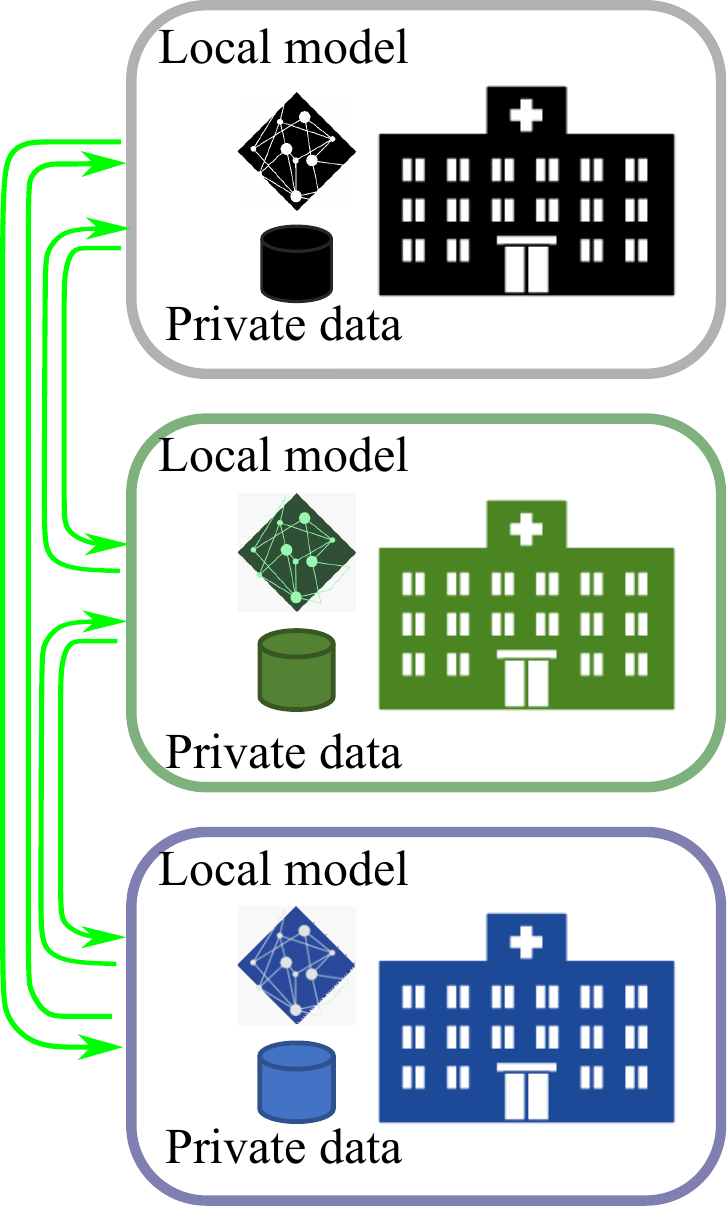}
\label{subfig:client-to-client}
}
\end{minipage}
\caption{The center-to-peer (a) and peer-to-peer (b) federated learning.}
\label{Fig:federatedLearning}
\end{figure}

To overcome the data privacy issue, federated learning has been proposed \cite{rieke2020future,sarma2021federated,dayan2021federated,xu2021federated,li2020federated},
 which enables multiple centers to train a high performance model without sharing data. In center-to-peer federated learning (C2PFL) (Fig.\,\ref{subfig:Center-to-client}), a central server is required to coordinate training information for a global model. However, such a central server is financially expensive to build and the communication cost between the central server and multiple clients is expensive as well \cite{xu2021federated}. 
  In IT industry, because of the extremely large number of mobile clients and the limited computational ability of mobile clients, C2PFL is preferred over peer-to-peer federated learning (P2PFL). However, for multicenter collaboration in medical fields, the number of centers is typically low and each center has sufficient computation resources. Therefore, P2PFL in a decentralised mode (Fig.\,\ref{subfig:client-to-client}) is more feasible in practice \cite{xu2021federated,wink2021approach}, since a central server is no longer required. The simplest way for such peer-to-peer federated learning is to continually train the same model one center after another via weight transfer (either \textbf{single weight transfer (SWT)} or \textbf{cyclic weight transfer (CWT)}) \cite{chang2018distributed,sheller2018multi,sheller2020federated}.


However, in P2PFL with naive weight transfer, performance fluctuation from center to center is observed \cite{chang2018distributed,sheller2018multi,sheller2020federated}. When a model is retrained on new datasets or tasks, deep learning suffers from the problem of catastrophic forgetting \cite{mccloskey1989catastrophic,lesort2020continual,delange2021continual}, i.e., deep learning models forget learned old knowledge catastrophically. To address the forgetting problem, continual learning has been proposed. According to application scenarios, three types of continual learning are investigated \cite{van2022three}: task-incremental learning \cite{delange2021continual}, class-incremental learning \cite{masana2020class}, and domain-incremental learning \cite{van2022three}. In the task-incremental learning, a model incrementally learns a set of distinct tasks, while in the class-incremental learning a model incrementally discriminates between a growing number of classes. Domain-incremental learning addresses the same task with the same number of classes, but the distribution of training/test data varies, e.g., cardiac image segmentation on heterogeneous data with domain discrepancy \cite{li2022domain}, where forgetting caused by domain shift needs to be addressed. Domain incremental learning is very similar to P2PFL for multicenter collaboration but with differences, e.g., the necessity of domain discrepancy.

So far, many continual learning algorithms have been proposed, which are mainly categorized to three categories \cite{delange2021continual}: replay methods \cite{isele2018selective,rolnick2019experience,chaudhry2019continual}, architectural (parameter isolation) methods \cite{rusu2016progressive,mallya2018packnet,wu2019large,fernando2017pathnet,aljundi2017expert,bang2021rainbow}, and regularization methods \cite{li2017learning,kirkpatrick2017overcoming,zenke2017continual}. Replay methods select representative samples from previous datasets to preserve learned knowledge. This is feasible to overcome data storage constraints, but not feasible for multicenter collaboration since samples from other centers are not available due to data privacy. In this case, using synthetic samples from generative adversarial networks (GANs) is an alternative choice \cite{shin2017continual}. Architectural methods design dynamic network architectures or dynamic parameters for multi-task scenarios, where certain parts of the network (e.g., certain modules \cite{aljundi2017expert,bang2021rainbow}, weights \cite{rusu2016progressive,mallya2018packnet,wu2019large}, or neuron connections \cite{fernando2017pathnet}) are responsible for certain tasks. Regularization methods use the same conventional neural networks, but with new regularization terms during training to preserve important parameters for learned knowledge, like \textbf{learning without forgetting (LWF)} \cite{li2017learning}, \textbf{elastic weight consolidation (EWC)} \cite{kirkpatrick2017overcoming} and \textbf{synaptic intelligence (SI)} \cite{zenke2017continual}. 

 To date, comprehensive surveys for task-incremental learning \cite{delange2021continual} and class-incremental learning \cite{masana2020class} have been conducted. However, domain-incremental learning is the least studied continual learning scenario, as stated in \cite{van2022three}, and its application to overcome data privacy in multicenter collaboration for medical applications has not been thoroughly investigated yet. The contributions of this work mainly lie in the following aspects:
\begin{itemize}

\item A comprehensive survey on the efficacy of different regularization-based continual learning methods for multicenter collaboration is performed, which can be sister surveys of \cite{delange2021continual} and \cite{masana2020class};

\item The influences of data heterogeneity, classifier head setting, network optimizer, model initialization, center order, and weight transfer type have been investigated thoroughly;

\item A framework to combine P2PFL and domain-incremental learning is established, which is publicly accessible\footnote{\url{https://github.com/YixingHuang/ITLsurvey}} to the research community for further development.
\end{itemize}

\section{Related Work}

This work is related to the work of P2PFL with naive weight transfer, federated continual learning, and domain-incremental learning.

P2PFL with weight transfer \cite{chang2018distributed,sheller2018multi,sheller2020federated} has been proposed for multicenter collaboration on medical applications. Chang et al. \cite{chang2018distributed} proposed distributing deep learning models via SWT or CWT for multi-institutional collaboration and the CWT method was shown to achieve comparable performance to that of centrally hosted patient data. Sheller et al. \cite{sheller2018multi,sheller2020federated} have applied the two weight transfer methods to glioma segmentation and confirmed that CWT is superior to SWT. However, the performance fluctuation is observed which is caused by forgetting \cite{mccloskey1989catastrophic} when the model is transferred from one institution to another. However, the forgetting problem has not been addressed.

 Recently, federated continual learning \cite{yao2020continual,yoon2021federated,usmanova2022federated,casado2022concept} becomes an emerging concept. However, these contributions were all proposed for C2PFL with different focuses, and hence are different from our peer-to-peer scenarios. Yao and Sun \cite{yao2020continual} proposed to estimate importance weights for the global model based on EWC and use such importance weights to constrain local training in clients to address the data heterogeneity problem. Yong et al. \cite{yoon2021federated} proposed to decompose model parameters into global parameters, local base parameters and task-adaptive parameters. Hence the global server can aggregate and redistribute information from different clients selectively, where continual learning is applied in local clients to reduce the communication cost. Usmanova et al. \cite{usmanova2022federated} proposed a federated learning without forgetting (FLwF) method for class-incremental scenarios \cite{masana2020class}, which uses the central server as the second teacher to guide a student client model to deal with catastrophic forgetting, along with the first teacher trained locally in each client. Casado et al. \cite{casado2022concept} addressed the concept drift problem in federated learning, where data feature distribution varies over time at each client. A simple rehearsal method which stores old memories at each local client is proposed for drift adaptation before global aggregation.

Domain adaptation is a common technique to address the performance drop in transfer learning \cite{niu2020decade}. However, typical domain adaptation techniques \cite{guan2021domain} aim to improve the model performance in the target domain while the performance on the source domain is no longer important. However, in multicenter collaboration, a global model with good performance on the test datasets from all centers is pursued. Domain incremental learning \cite{van2022three} aims to address the domain discrepancy problem and preserve the model performance on data from different domains using continual learning techniques. Its sequential training scheme fits well to P2PFL. Many domain incremental learning algorithms have been integrated into medical applications \cite{wachinger2020importance,li2022domain,you2022incremental,xie2022general,huang2022continual,srivastava2021continual,thakur2023self}. For example, Srivastava et al. proposed a  vector quantization memory buffer to efficiently store and replay hidden representations of data from different domains \cite{srivastava2021continual}. Li et al. proposed a domain incremental method with style-oriented replay and domain sensitive feature whitening for cardiac image segmentation \cite{li2022domain}. But the domain identity information is unknown for the training and test data in their setting, which is different for P2PFL in multicenter collaboration. Thakur et al. \cite{thakur2023self} proposed a replay-based self-aware stochastic gradient (SGD) method for domain shifts over time in a fixed center instead of multicenters. You et al. used an encoder to learn domain discrepancy between the current dataset and old exemplars in an \textbf{incremental transfer learning (ITL)} setting \cite{you2022incremental}, but the old exemplars typically are challenging to obtain in multicenter collaboration because of data privacy regulations. In our previous preliminary work \cite{huang2022continual}, SI was applied to address the forgetting problem in multicenter collaboration for brain metastases segmentation, where many questions remain unresolved, which lead to this comprehensive survey.

\section{Problem and Learning Framework}

\subsection{Problem Definition}
Conventional continual learning topics typically focus on addressing the task-incremental learning \cite{delange2021continual}, and class-incremental learning \cite{masana2020class} problems. P2PFL for multicenter collaboration is very similar to domain incremental learning \cite{van2022three} but without the necessity of domain discrepancy. For multicenter collaboration in this work, learning for a single task with a fixed number of classes but incremental data from multiple centers is addressed. During the training and test stages, the training and test data is located locally in each center and hence the center identity information is known. The deep learning model iteratively learns from a sequence of datasets $\left\lbrace \mathcal{C}^{(1)}, \mathcal{C}^{(2)},..., \mathcal{C}^{(N)}\right\rbrace$ where $\mathcal{C}^{(\mu)}$ is a labeled dataset from the $\mu$-th center, $\mathcal{C}^{(\mu)} = \left\lbrace \vx_i^{(\mu)}, \vy_i^{(\mu)}\right\rbrace_{i=1}^{N_{\mu}}$, which consists of $N_{{\mu}}$ pairs of instances $\vx_i^{(\mu)}$ and their corresponding labels $\vy_i^{(\mu)}$. Considering data privacy constraints, center $\mu$ has the access to its own data $\mathcal{C}^{(\mu)}$ for training a shared model and the datasets from other centers are all inaccessible. Given $\mathcal{C}^{(\mu)}$ and a model trained in a previous center, the primary learning objective at the $\mu$-th center is to minimize the task-specific loss $\mathcal{L}_{\text{task}}$,
\begin{equation}
\arg\min_{\vtheta^{(\mu)}} \mathcal{L}_{\text{task}}\left(\vtheta^{(\mu)}; \vtheta^{(\mu-1)}, \mathcal{C}^{(\mu)}\right),
\label{eqn:taskLoss}
\end{equation}
where $\vtheta^{(\mu)}$ is the parameter set of the model at the $\mu$-th center, while the model performance on test data in previous centers is desired to be preserved. Due to the incremental weight transfer nature of our problem setting, we call it \textbf{incremental transfer learning (ITL)} named after \cite{you2022incremental}, but without the use of old exemplars because of data privacy.

The problem definition of our ITL is very similar to domain-incremental learning \cite{van2022three} but with the following differences:
\begin{itemize}
\item In domain-incremental learning datasets must be from different domains, while in multicenter collaboration, datasets from different centers can be from the same domain, because the data can be collected from the same medical imaging devices with the same preprocessing pipeline.

\item Mixing representative sample data together with new training data is possible for general domain-incremental learning, while it is impossible in ITL for multicenter collaboration because of data privacy. However, the model can revisit each center via CWT for training and test in ITL for multicenter collaboration.

\item Domain-incremental learning typically aims at domain-specific (center-specific in this work) output instead of a global output (Fig. 2 in \cite{van2022three}). Therefore, multi-head setting is typically used for optimal output at each domain/center. However, for multicenter collaboration, a global model exhibiting generalizability across datasets from all the centers is pursued.

\item For domain incremental learning, the source information (domain/center) of the training or test data might be unknown \cite{li2022domain}, while it is clearly known for multicenter collaboration.
\end{itemize}

\subsection{Regularization methods for continual learning}
As direct optimization based on the primary task-specific loss (Eqn.\,(\ref{eqn:taskLoss})) alone has the risk of forgetting, continual learning techniques are applied. For continual learning, architectural methods are not our choice since they are mainly designed for multi-task scenarios instead of the single-task scenario in multicenter collaboration. Replay methods require samples from other centers, which is typically not easily feasible due to data privacy constraints. Generative replay methods using synthetic samples or extracted representative features is a promising approach. Nevertheless, in this work, we focus on regularization methods and the following classic regularization-based continual learning techniques are investigated: LWF \cite{li2017learning}, EWC \cite{kirkpatrick2017overcoming}, SI \cite{zenke2017continual}, \textbf{memory aware synapse (MAS)} \cite{aljundi2018memory}, \textbf{encoder based lifelong learning (EBLL)} \cite{rannen2017encoder}, and \textbf{incremental moment matching (IMM)} \cite{lee2017overcoming}. The overall loss function $L_{\mu}$ for center $\mu$ for these regularization methods can be represented as follows in general,
\begin{equation}
\mathcal{L}\muindex = \mathcal{L}_{\textrm{task}} + \lambda\muindex \phi\left(\boldsymbol{\theta}\muindexpre, \boldsymbol{\theta}\right),
\label{eq:regularizationLoss}
\end{equation}
where $\boldsymbol{\theta}$ is the current parameter set to optimize during training, while $\boldsymbol{\theta}^{\mu-1}$ is the parameter set transferred from the previous center, $\mu$ is the index for the current center, and $\lambda$ is a regularization parameter to control the amount of the regularization term $\phi$. 

EWC \cite{kirkpatrick2017overcoming}, SI \cite{zenke2017continual}, and MAS \cite{aljundi2018memory} share the same idea of penalizing the change of important network parameters, but they have different ways to compute the importance weights $\boldsymbol{\Omega}^{\mu}$: EWC computes the importance weights after model training  using the Fisher information matrix; SI computes the importance weights during training by tracking the magnitudes of parameter changes; MAS computes the importance weights after training by measuring the change of the model output with respect to the parameter changes. The overall loss function $L_{\mu}$ of the $\mu$-th center for these three methods can be represented as follows in general,
\begin{equation}
\mathcal{L}\muindex = \mathcal{L}_{\textrm{task}} + \lambda\muindex \sum_k \boldsymbol{\Omega}\muindexpre_k \left(\boldsymbol{\theta}\muindexpre_{k} - \boldsymbol{\theta}_k\right)^2,
\label{eq:SIcontinualLoss}
\end{equation}
where the right term penalizes the change of important network parameters trained in previous centers. $\lambda$ is the relaxation parameter to trade off old versus new knowledge and $k$ is the index for a certain parameter.

LwF \cite{li2017learning} controls forgetting by imposing network output stability. In other words, for certain samples, the model after training with local data should get similar predictions to those before training. The objective function for LwF is as follows,
\begin{equation}
\begin{array}{l}
\mathcal{L}\muindex_{\text{LwF}} = \mathcal{L}_{\textrm{task}} + \\
\qquad \qquad\lambda\muindex \mathcal{L}_{\textrm{KDL}}\left(\mathcal{M}({\mathcal{C}}\muindex, \boldsymbol{\theta}\muindexpre),  \mathcal{M}\left({\mathcal{C}}\muindex, \boldsymbol{\theta}\right)\right),
\end{array}
\label{eq:LwFLoss}
\end{equation}
where $\mathcal{L}_{\textrm{KDL}}$ is the knowledge distillation loss (KDL) \cite{hinton2015distilling} and $\mathcal{M}$ is the network model. In ITL, samples from the previous center typically is not available. Therefore, the training dataset of ${\mathcal{C}}\muindex$ is used as the sample set for computing the KDL.

EBLL \cite{rannen2017encoder} extends LwF by using an autoencoder to capture important task-specific features in a low dimensional manifold. The main network $\mathcal{M}$ is divided to two parts: feature extractor $\mathcal{F}$ and task/center-specific classifier $\mathcal{T}$. An autoencoder is trained to project features extracted by $\mathcal{F}$ onto a lower dimensional manifold and reconstruct them back for each task.
With the autoencoder trained from the previous center, the following objective function is used for training the main network in the $\mu$-th center,
\begin{equation}
\begin{array}{l}
\mathcal{L}\muindex = \mathcal{L}\muindex_{\text{LwF}} + \\
\alpha/2\lVert\mathcal{E}\muindexpre\left(\mathcal{F}\left(\vx\muindex,\boldsymbol{\theta}\right)\right) - \mathcal{E}\muindexpre\left(\mathcal{F}\left(\vx\muindex,\boldsymbol{\theta}\muindexpre\right)\right) \rVert_2^2
\end{array}
\label{eq:EBLLloss}
\end{equation}
where $\mathcal{E}\muindexpre$ is the encoder part of the autoencoder trained in the previous center and $\alpha$ is a parameter for the additional constraint on the manifold space.

IMM \cite{lee2017overcoming} merges models trained from each center to obtain an improved model.
 Such model merging relies on a flat or convex loss surface, which is obtained by weight transfer, L2 transfer and drop transfer. The L2 transfer uses an $\ell_2$ regularization to get a smooth loss surface between $\boldsymbol{\theta}^{\mu-1}$ and $\boldsymbol{\theta}^{\mu}$,
 \begin{equation}
 \mathcal{L}\muindex = \mathcal{L}_{\textrm{task}} + \beta \left(\boldsymbol{\theta}\muindexpre - \boldsymbol{\theta}\right)^2,
\label{eq:L2Transfer}
 \end{equation}
 where $\beta$ is a small value (e.g. 0.001). After training in all the centers, the models are merged as the following,
 \begin{equation}
\hat{\boldsymbol{\theta}}_k\muindex = \sum_{\nu=1}^{\mu}\alpha^{(\nu)}\boldsymbol{\theta}_k^{(\nu)},
\end{equation}
where $\alpha^\nu$ sums up to 1 for all centers. Two ways for model merging have been proposed in \cite{lee2017overcoming}: mean-IMM and mode-IMM. Mean-IMM minimizes the KL-divergence between a model trained from an individual task (center) and the model trained from all tasks (mix data), and $\alpha^\nu$ simply equals $1/N$, where $N$ is the total number of centers. Mode-IMM merges model parameters using a Laplacian approximation to estimate modes (similar to local minima), and $\alpha^\nu$ is determined by parameter diagonal covariance matrices. One way to compute $\alpha^\nu$ is to use the inverse of a Fisher information matrix as an indicator of each parameter's importance $\boldsymbol{\Omega}^{\mu}_k$ like the EWC method \cite{kirkpatrick2017overcoming},
\begin{equation}
\alpha\muindex = \boldsymbol{\Omega}\muindex_k/\sum_{\nu=1}^\mu\boldsymbol{\Omega}^{(\nu)}_k.
\end{equation}

\subsection{Incremental transfer learning framework}
The multicenter collaboration scenario is different from the typical continual learning scenarios \cite{van2022three}. Hence, the direct application of existing continual learning frameworks is suboptimal for multicenter collaboration. Therefore, the following components are suggested for our ITL framework.

\subsubsection{Single-head instead of multi-head setting}
For task-incremental learning and class-incremental learning, the multi-head setting imposes a separate classifier for each task to achieve best adaptability. Especially, because of the different number of classes, the old classifiers cannot be used for new tasks. For domain-incremental learning, although the same task is optimized, the multi-head setting is typically used for optimal output at each context/center \cite{van2022three}. However, when a trained classifier $\mathcal{T}$ is replaced by a new classifier $\mathcal{T}'$ for training in the current center, the model parameters in the feature extractor $\mathcal{F}$ will also be updated to a new one $\mathcal{F}'$. Although the combination of  $\mathcal{F}'$ and $\mathcal{T}'$ is optimal for the current center, feature mismatch will occur between the updated feature extractor $\mathcal{F}'$ and the classifier for the previous center $\mathcal{T}$. Therefore, catastrophic forgetting occurs. To avoid such mismatch, the single-head setting is used in our proposed ITL framework so that the feature extractor and the classifier are always optimized jointly.

\subsubsection{Reload optimizers}
For task-incremental continual learning, maximum plasticity is achieved if the model is trained solely on the dataset for this certain task. Therefore, a learning rate grid search (LRGS) step \cite{li2020federated,xu2021federated} is commonly used to achieve the maximum plasticity for new tasks and a new training with the obtained learning rate starts after weight transfer. However, for multicenter collaboration, due to the limited amount of data at each center, training solely on the local data will not lead to maximum plasticity. Therefore, such a LRGS step is not necessary. In fact, a sudden change of learning rate will lead to performance fluctuations for other centers. Therefore, a smooth learning rate transition from center to center is preferred. 
For this purpose, a simple solution is simply reloading optimizers from the previous center for the training in the current center, while LRGS is used only in the first center.

\subsubsection{Use adaptive optimizers}
So far, all the above continual learning methods use SGD for optimization \cite{li2017learning,aljundi2018memory,kirkpatrick2017overcoming,zenke2017continual,lee2017overcoming,rannen2017encoder,delange2021continual,masana2020class}. SGD achieves good performance in tasks. However, the use of SGD is known to be difficult for initial training and needs a proper learning rate scheduler for model convergence. Adaptive optimizers like adaptive adaptive moment estimation (Adam) automatically adjust the learning weight for each model parameter. Hence, they will relieve the choice of learning rate schedulers. To integrate the regularization-based continual learning techniques, the adaptive optimizers need to be modified. An example of using Adam for SI is presented in Appendix \ref{subsect:SI_Adam}.

\subsubsection{Overfitting monitor}
To avoid overfitting to local data, the learning rate is decayed if no performance gain is achieved after every $E_\text{val}$ epochs at each center. Early stopping is used if no performance gain is achieved after $E_\text{stop}$ epochs. The model with the best validation loss is shared to the next center. Such monitoring for overfitting reduces the overfitting problem, although overfitting typically still exists. 

\subsubsection{Cyclic weight transfer}
\begin{figure}
\centering
\includegraphics[width=\linewidth]{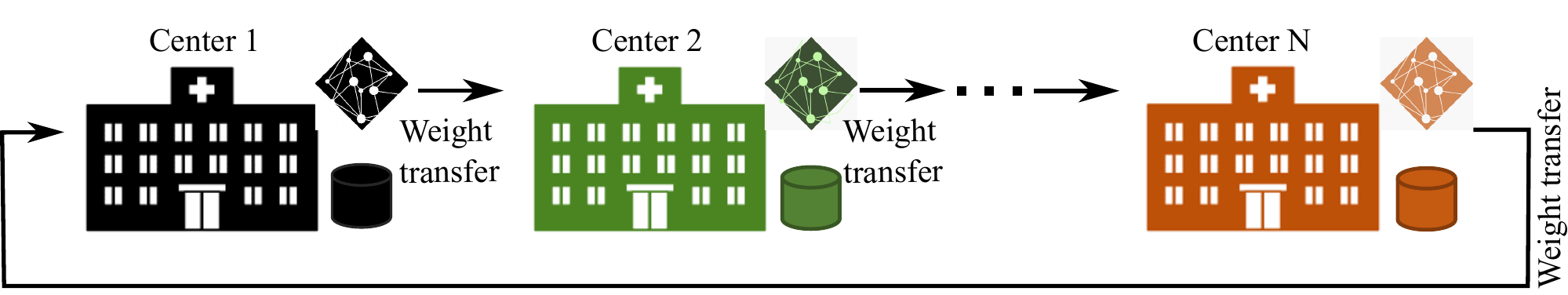}
\caption{Cyclic weight transfer (CWT) for multicenter collaboration, where the weights of a trained model and its optimizer at each center are transferred from Center 1 to Center N successively for continual training.}
\label{Fig:CyclicWeightTransfer}
\end{figure}

For continual learning, a common setting is that the datasets from previous tasks are no longer accessible. Therefore, SWT instead of CWT is applied. For ITL, datasets from different centers can be re-accessed via CWT and CWT has been reported to achieve better performance than SWT \cite{chang2018distributed,sheller2018multi,sheller2020federated}. Therefore, CWT is integrated with continual learning techniques in our ITL framework, as displayed in Fig.\,\ref{Fig:CyclicWeightTransfer}.

With the above ITL framework, the efficacy of different continual learning regularization methods is investigated on different applications.

\section{Experiments on Tiny ImageNet Dataset}
\subsection{Experimental Setup}
\subsubsection{Dataset}
As a proof-of-concept investigation, image classification experiments on the Tiny ImageNet dataset \cite{le2015tiny} in the multicenter collaboration scenario are conducted. The image size is $64 \times 64$. 10 classes among the 200 classes from the Tiny ImageNet dataset are chosen. Each class contains 500 samples subdivided into training (80\%) and validation (20\%), and
50 samples for test. These images are equally split into 5 centers, i.e., each center 80 images per class (in total 800 images) for training, 20 images per class (in total 200 images) for validation and 10 images per class (in total 100 images) for test. As the original Tiny ImageNet images are randomly split into each center, the data distribution among all the 5 centers can be assumed to be \textbf{independent and identically distributed (IID)}.
 
 In practice, data from different centers might be heterogeneous. To simulate non-IID data, Gaussian noise with a zero mean and a standard deviation of 25 is added to the images in one center. The noise is first added to the fifth center and then the datasets from the third and fifth centers are swapped to investigate the influence of training order.
 
\subsubsection{Network}
The small VGG-9 network from \cite{delange2021continual} is used, which has 6 convolutional layers (first 4 layers with 64 filters and the remaining 2 layers with 128 filters), 4 max-pooling layers and 3 fully connected layers (128 classifier dimensions). The task-specific loss $\mathcal{L}_{\textrm{task}}$ is the cross-entropy loss.

For SWT, all the models are trained for 50 epochs at each center. For CWT, a weight transfer frequency of $E_\text{transfer}=10$ is investigated with a total number of 5 iterations. For overfitting monitor, the learning rate is decayed into half after every $E_\text{val} = 5$ epochs and early stopping is executed after $E_\text{stop} = 20$ epochs. The joint model is trained from mixed data for 50 epochs. A constant $\lambda$ value of 1 for LwF, EWC, SI and MAS is used. For IMM, the $\lambda$ value for the L2-weight transfer is 0.01. The batch size is set to 100. The Adam optimizer as a representative of adaptive optimizers is investigated in comparison to the widely-used SGD optimizer in continual learning. For SGD, the initial learning rate is set to 0.1 after LRGS. An exponentially decaying learning rate scheduler is set for the SGD optimizer, $r = 0.1 *0.8^{e/5}$, where $e$ is the accumulated epoch number. For Adam, the learning rate is fixed to 0.001 after LRGS.

\subsubsection{Evaluation metrics}
 
\

\textbf{Accuracy:} For the test data in the $\mu$-th center, an accuracy sequence $\boldsymbol{a}\muindex = a\muindex_1, a\muindex_2, ...,a\muindex_{N_\text{total}}$ is computed for a sequence of trained models, where $N_\text{total}$ is the total number of center-wise training. The average accuracy $a_\text{mean}$ of the final model evaluated in all the centers is reported for each method,
\begin{equation}
a_\text{mean} = \sum_{\mu=1}^{N}a\muindex_{N_\text{total}}/N.
\end{equation}
With the single-head setting, forgetting is no longer catastrophic. Therefore, the forgetting metric in \cite{delange2021continual} is not reported in this work.

\textbf{Monotonicity:} As a monotonic increase of accuracy is desired, a monotonicity sequence is computed as well, $\boldsymbol{m}\muindex = m\muindex_2, m\muindex_3, ..., m\muindex_{N_\text{total}}$, where $m\muindex_i = 1 \text{ if } a\muindex_i \geq a\muindex_{i-1} \text{ else } 0$. For each method, the average monitonicity of all the models is,
 \begin{equation}
 m_\text{mean} = \sum_{\mu=1}^{N}\sum_{i=2}^{N_\text{total}}m\muindex_{i}/(N\cdot (N_\text{total}-1)),
 \end{equation}
where $m_\text{mean}$ is in the range of [0, 1]. $m_\text{mean} = 1$ stands for fully monotonic increase and $m_\text{mean} = 0$ stands for fully monotonic decrease, while $m_\text{mean} = 0.5$ stands for 50\% of weight transfers leading to accuracy increase (the most oscillation situation). 

\textbf{Significance:} Due to the stochastic nature of model training, each method is repeated 30 times with different random seeds. The mean accuracy, standard deviation of accuracy, and mean monotonicity are displayed. In addition, the statistical significance ($p < 0.05$) of each method compared with FT is reported: ``Yes+" indicates whether a method is superior to FT with significance; ``Yes-" indicates whether a method is inferior to FT with significance; otherwise, ``No" is displayed to indicate no statistical significance.

\subsection{Results of IID data}
\begin{figure}
\centering
\includegraphics[width=\linewidth]{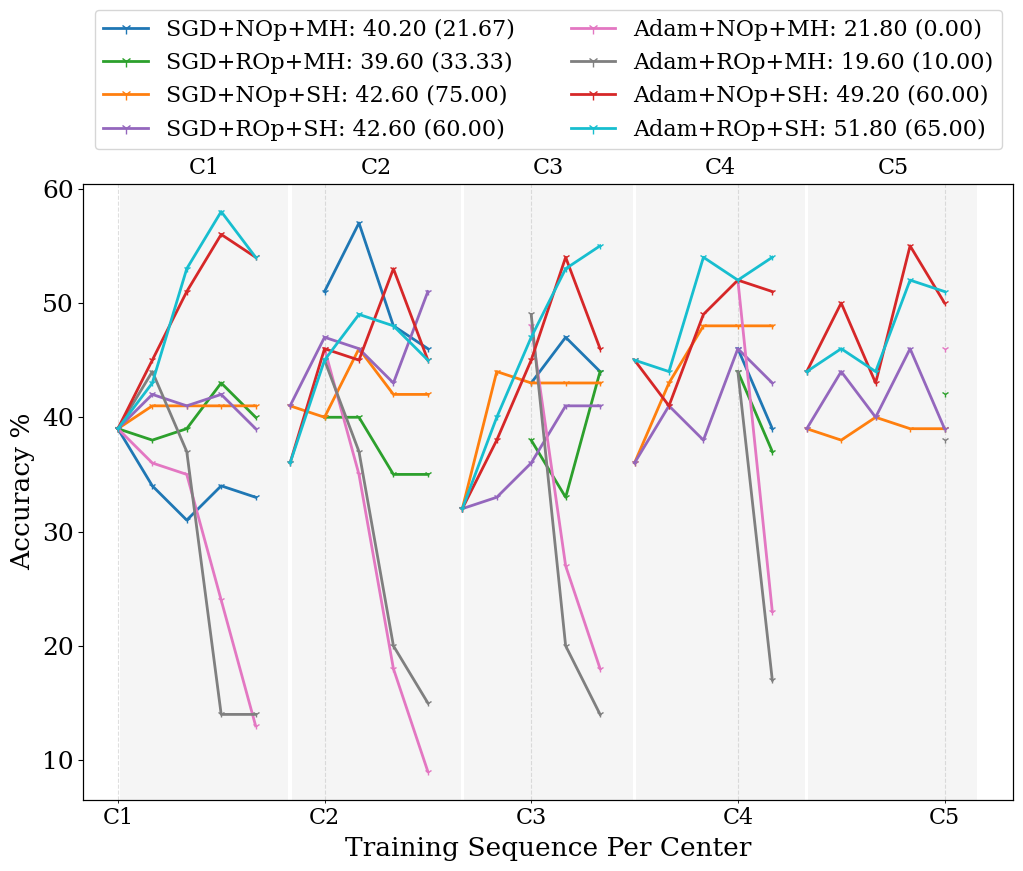}
\caption{The naive SWT (i.e., FT) performance with different configurations on IID data. The configurations include SGD vs. Adam, new optimizer (NOp) vs. reload optimizer (ROp), and multi-head (MH) vs. single-head (SH).  The average accuracy (monotonicity) for each method is reported in the legend. The implementation and performance visualization are adapted from \cite{delange2021continual}. }
\label{Fig:FTSettings}
\end{figure}
\subsubsection{\textbf{Improve the baseline configuration for naive SWT}}
We first investigate the performance of naive SWT, i.e. \textbf{fine-tuning (FT)} after weight transfer, on IID data using different configurations. When the network uses the multi-head setting, since no classifier heads are available for the subsequent centers, the trained model is only shared to the previous centers for evaluation to track the performance variation, as displayed in Fig.\,\ref{Fig:FTSettings}. When the single-head setting is used, each trained model is shared to all the centers for evaluation.

The basic configuration for domain-incremental learning scenarios is SGD + new optimizer (NOp) + multi-head (MH). In this setting, apparent forgetting is observed as the accuracy curves drop in multiple sites. With the SGD optimizer, both reloading optimizer (ROp) and single-head (SH) have a minor impact on the overall accuracies, all staying around 40\%.

 With the Adam optimizer, the multi-head setting leads to catastrophic forgetting of the trained models, since FT only has low accuracies of 21.80\% and 19.60\% for NOp and ROp, respectively. The forgetting problem is more severe for Adam than SGD with the multi-head setting, because Adam allows a larger learning rate for training.
 When the single-head (SH) setting is used, FT achieves high accuracies of 49.20\% and 51.80\% for NOp and ROp, respectively. Since the configuration of Adam + ROp + SH achieves the best accuracy among different configurations (also better than all the configurations with SGD), it is used as the baseline configuration to investigate the influence of continual learning regularization methods.

In our experiments, when the SGD optimizer is replaced by Adam, the accuracies of \textbf{joint} training (all the data from different centers is mixed together) and \textbf{independent training (IT)} (each center trains a model independently from its own data) are improved in general as well. This confirms the benefit of using an adaptive optimizer.

\begin{figure}
\centering
\includegraphics[width=\linewidth]{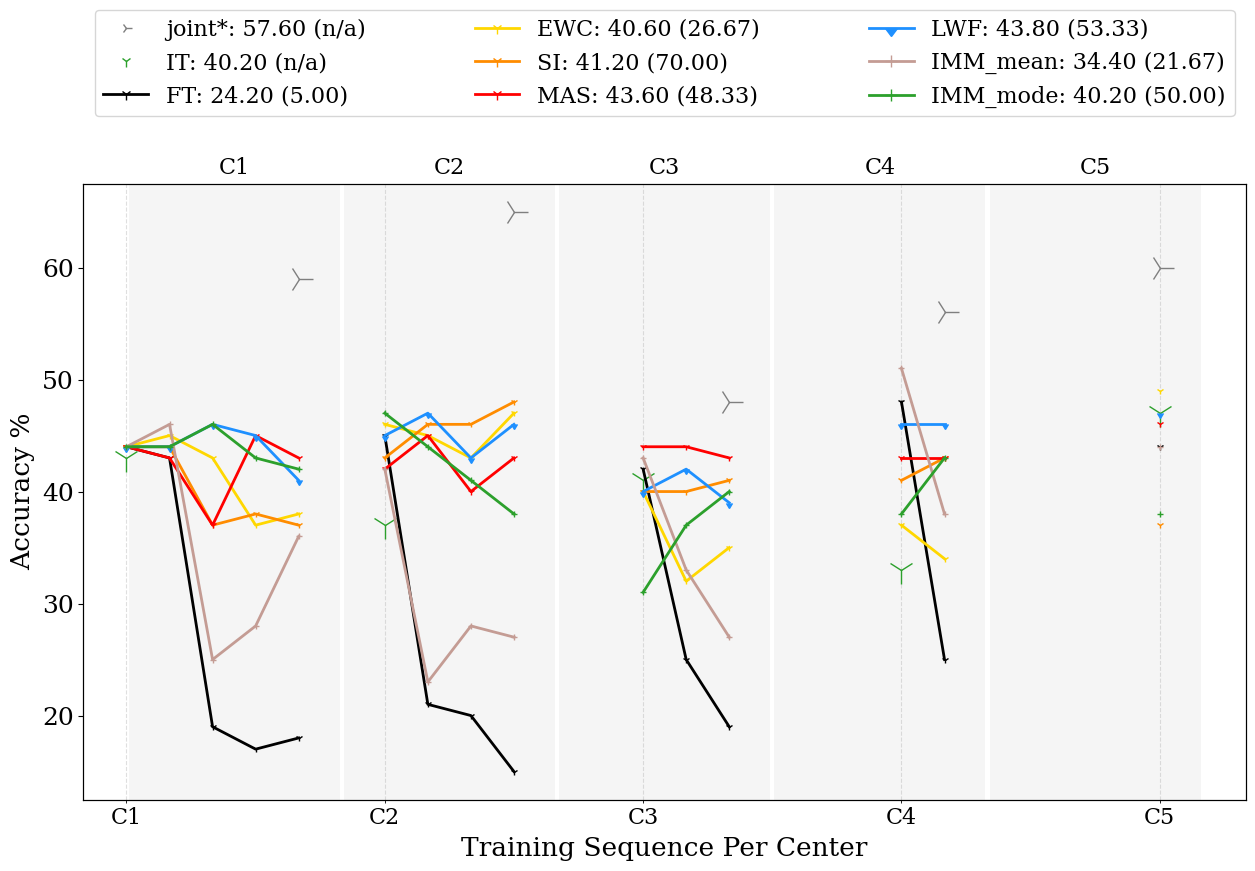}
\caption{The SWT results of continual learning regularization methods on the Tiny ImageNet data with the setting: Adam + new optimizer (NOp) + multi-head (MH).}
\label{Fig:baseTICLAdam}
\end{figure}

\subsubsection{\textbf{Continual learning regularization avoids catastrophic forgetting in the multi-head setting}}
When the typical configuration for continual learning is used for ITL but with the Adam optimizer (i.e., Adam + NOp + MH), the accuracy plots of different methods are displayed in Fig.\,\ref{Fig:baseTICLAdam}.
The accuracy of FT for Center 1 stays around 41\% after training in Center 2. However, it decreases drastically to below 20\% after the subsequent trainings. Its accuracies for other centers drop drastically as well after training in the next centers. 
This is caused by the large change in the feature extractor ($\mathcal{F}$) parameters after training in different centers, which leads to the mismatch of the feature extractor $\mathcal{F}$ and the center-specific classifier $\mathcal{T}$. 

Instead, all the continual learning regularization methods, except IMM-mean, achieve relatively stable accuracies. Although the accuracy of IMM-mean for Center 1 drops together with that of FT first, later it turns to rise back. 
It is worth mentioning that SI and FT have the same implementation in our work except that $\lambda$ is set to zero for FT in Eqn.\,(\ref{eq:regularizationLoss}). Fig.\,\ref{Fig:baseTICLAdam} demonstrates that all the continual learning regularization methods are effective to reduce the catastrophic forgetting problem for single-task multicenter collaboration with the multi-head setting. 

\subsubsection{\textbf{Continual learning regularization methods improve monotonicity but not accuracy for IID data}}

The performances of different methods with SWT for IID data are displayed in Tab.\,\ref{Tab:SWT_IID}. With all random initializations in the first center (for 30 repeats, all the models trained in the first center use different random seeds), the average accuracy of FT is 49.49\%, while those of joint and IT are 57.30\% and 39.26\%, respectively. Among all the continual learning regularization methods, EWC, SI and MAS have similar or even worse monotonicity, but LWF, EBLL, IMM-mean and IMM-mode have higher monotinicity than FT. Nevertheless, all the continual learning regularization methods cannot improve the accuracy, and MAS, EBLL and IMM-mode even have significantly worse accuracies. The same conclusion is observed from the CWT results for IID data displayed in Tab.\,\ref{Tab:CWT_IID}, as all the continual learning regularization methods are not significantly better than FT when the initial model uses 10 epochs, although the monotonicity of LWF, EBLL, IMM-mean and IMM-mode is better.

\subsubsection{\textbf{Initialization model matters for SWT}}
The 30 models trained in the first center have an average accuracy of 40.90\%. Among them, a low accuracy model (35.0\%) and a high accuracy model (45.2\%) are chosen as a fixed initial model for subsequent SWT training. In general, all methods achieve higher accuracies when a high-accuracy model is used as initialization than a low-accuracy model. Especially, with non-IID data (noisy data in Center 5) as displayed in Tab.\,\ref{Tab:SWT_N5}, LWF, EBLL, IMM-mean and IMM-mode are significantly superior to FT when a high-accuracy initialization is used, while EBLL and IMM-mean are significantly worse than FT when a low-accuracy initialization is used.
For LWF and EBLL, the model trained in the first center is used as a teacher model for the training in the second center. Therefore, a high accuracy is beneficial to improve the accuracies of subsequent models. For IMM-mean and IMM-mode, the model weights trained in the subsequent centers are all constrained  (by the L2 transfer) to be close to the weights in the first center and the model trained in the first center always contributes to the subsequent merged models. Therefore, a high performance initialization is crucial for the performance of the final merged model.

\subsubsection{\textbf{Non-convergent model is preferred for initialization in CWT}}
The CWT results on IID data are displayed in Tab.\,\ref{Tab:CWT_IID}. When the initial model is trained with 10 epochs in Center 1, FT achieves an average accuracy of 56.18\%, which is slightly lower than (0.82\% with significance) but very close to that of joint data. All other methods except MAS also achieve very high accuracies. In comparison, when the initial model is trained with 50 epochs in Center 1 (models typically converge before 50 epochs), most methods (except IMM-mean and IMM-mode without significance) achieve a significantly lower average accuracy, irrespective if a high-accuracy initial model or a low-accuracy initial model is used, on both IID and non-IID data. Please see the CWT results with convergent models as initializations in Tab.\,\ref{Tab:CWT_IID_App} and Tab.\,\ref{Tab:CWT_N5_App} in the appendix. 
A convergent model in the first center very likely reaches a local minimum, which has high energy barriers to reach other local minima or the global minimum. Hence, it is more difficult to be optimized to reach a better minimum in the subsequent training than a non-convergent model. Therefore, a non-convergent model is preferred for the initialization in CWT, which is in a stark contrast to the observation that SWT prefers a high-performance initial model.

%

\begin{table*}
\centering
\caption{The average performances of different methods with SWT for IID Tiny ImageNet data (\%)(30 repeats).}
\label{Tab:SWT_IID}
\begin{tabular}{l|l|c|c|c|c|c|c|c|c|c|c}
\hline
Initialization & Metric & joint & IT & FT & EWC &SI  & MAS & LWF & EBLL & IMM-mean & IMM-mode\\
\hline
\multirow{4}{*}{Random (40.90)} & accuracy& 57.30& 39.26& 49.49 & 49.62 & 49.12 & 44.25 & 47.96 &46.36 & 48.12 & 47.35\\
& standard deviation &1.54 & 2.26 & 2.59 & 2.45 & 2.42 & 2.05 & 2.15 & 1.64 & 2.40 & 2.13\\
&monotonicity& n.a.&n.a. & 71.03 & 70.69 & 68.28 & 65.86 & \textbf{78.79} &\textbf{77.93} & \textbf{80.17} & \textbf{74.31}\\
&significance & n.a. & n.a. &n.a.&No & No & Yes- &No & Yes- & No & Yes-\\
\hline
\multirow{4}{*}{Low (35.0)} & accuracy & n.a.& n.a.&  49.07& 49.1 & 48.84 & 42.97 & 49.19 &45.65& 46.63 & 48.14\\
&standard deviation& n.a. & n.a.&  1.94& 1.62 & 1.43 & 0.88 & 1.42 &2.13 & 1.53 & 1.52\\
&monotonicity& n.a. & n.a.&  78.83& 81.5 & 80.67& 78.0& \textbf{86.83} &\textbf{85.33} & \textbf{88.0} & \textbf{86.5}\\
&significance & n.a. & n.a. &n.a.&No & No & Yes- &No & Yes- & Yes- & No\\
 \hline
 \multirow{4}{*}{High (45.2)}&accuracy &n.a. & n.a. & 51.44 & 51.71 & 51.19 & 47.53& 50.43 & 48.73 & 51.1& 49.07\\
& standard deviation &n.a. & n.a.&1.86 & 1.73 & 1.55 & 1.26 & 1.12 &1.59 & 0.89 & 1.22\\
&monotonicity&n.a. & n.a.& 67.0 & 68.83 & 67.5 & 63.33 & \textbf{74.5} &\textbf{73.5} & \textbf{73.0} & 67.0\\
&significance & n.a. & n.a. &n.a.&No & No & Yes- &No & Yes- & No & Yes-\\
 \hline
\end{tabular}
\end{table*}

\begin{table*}
\centering
\caption{The average performances of different methods with CWT for IID Tiny ImageNet data (\%)(30 repeats).}
\label{Tab:CWT_IID}
\begin{tabular}{l|l|c|c|c|c|c|c|c|c|c|c}
\hline
Initialization & Metric & joint & IT & FT & EWC &SI  & MAS & LWF & EBLL & IMM-mean & IMM-mode\\
\hline
\multirow{2}{*}{Random (27.25)} & accuracy & 57.30& 39.26 &\textbf{56.18} & \textbf{56.03} & \textbf{55.47} & \textbf{50.82} & \textbf{56.33} & \textbf{56.62} &  \textbf{55.34} &  \textbf{55.23}\\
& standard deviation &1.54& 2.26&1.54 & 1.72 & 1.22 & 3.56 & 1.89&2.07 & 2.44 & 2.34\\
\multirow{2}{*}{(10 epochs)}&monotonicity& n.a. & n.a.&  63.19& 63.0 & 63.86& 64.67 & \textbf{71.73} &\textbf{72.63} & \textbf{75.10} & \textbf{74.30}\\
&significance & n.a. & n.a. &n.a.&No & No & Yes- &No & No & No & No\\
\hline
\end{tabular}
\end{table*}

\begin{table*}
\centering
\caption{The average performances of different methods with SWT for non-IID Tiny ImageNet data (noisy data in Center 5) (\%)(30 repeats).}
\label{Tab:SWT_N5}
\begin{tabular}{l|l|c|c|c|c|c|c|c|c|c|c}
\hline
Initialization & Metric & joint & IT & FT & EWC &SI  & MAS & LWF & EBLL & IMM-mean & IMM-mode\\
\hline
\multirow{4}{*}{Random (40.75)} & accuracy& 55.62& 39.12& 46.51 & 46.75 & 46.13 & 42.47 & \textbf{48.21} &46.97 & 47.84 & 47.78\\
& standard deviation &1.42 & 1.94 & 2.26 & 1.41 & 1.88 & 2.26 & 1.83 & 1.99 & 1.98 & 2.02\\
&monotonicity& n.a.&n.a. & 61.17 & 62.67 & 61.83 & 57.83 & \textbf{77.5} &\textbf{74.33} & \textbf{74.83} & \textbf{75.17}\\
&significance & n.a. & n.a. &n.a.&No & No & Yes- &\textbf{Yes+} & No & No & No\\
\hline
\multirow{4}{*}{Low (34.4)} & accuracy & n.a.& n.a.&  47.07& 46.26 & 46.03 & 38.52 & 47.52 &44.53 & 44.82 & 46.11\\
&standard deviation& n.a. & n.a.&  1.73& 1.92 & 1.58 & 1.06 & 1.58 &1.99 & 2.19 & 2.01\\
&monotonicity& n.a. & n.a.&  74.17& 74.33 & 73.0& 63.0 & \textbf{84.17} &\textbf{80.83} & \textbf{88.17} & \textbf{85.33}\\
&significance & n.a. & n.a. &n.a.&No & No & Yes- &No & Yes- & Yes- & No\\
 \hline
 \multirow{4}{*}{High (43.8)}&accuracy &n.a. & n.a. & 47.11 & 46.67 & 46.97 & 45.69 & \textbf{49.91} & \textbf{48.55} & \textbf{50.67} & \textbf{48.24}\\
& standard deviation &n.a. & n.a.&1.57 & 1.93 & 1.44 & 1.46 & 1.16 &1.91 & 1.08 & 1.53\\
&monotonicity&n.a. & n.a.& 54.83 & 55.67 & 56.83 & 56.0 & \textbf{72.0} &\textbf{71.17} & \textbf{71.5} & \textbf{70.17}\\
&significance & n.a. & n.a. &n.a.&No & No & Yes- &\textbf{Yes+} & \textbf{Yes+} & \textbf{Yes+} & \textbf{Yes+}\\
\hline
 \multirow{2}{*}{High (43.8)}&accuracy &n.a. & n.a. & 46.89 & 46.99 & 47.53 & 46.51 & 47.58 & 47.26 & \textbf{49.25} & \textbf{48.11}\\
& standard deviation &n.a. & n.a.&1.60 & 1.45 & 1.87 & 1.39 & 1.28 &1.50 & 1.24 & 1.66\\
 \multirow{2}{*}{4-center training} &monotonicity &n.a. & n.a.& 58.0 & 61.56 & 63.78 & 61.33& 67.78 &69.78 & 70.44 & 71.33\\
&significance & n.a. & n.a. &n.a.&No & No & No &No & No & \textbf{Yes+} & \textbf{Yes+}\\
 \hline
\end{tabular}
\end{table*}

\begin{table*}
\centering
\caption{The average performances of different methods with SWT for non-IID Tiny ImageNet data (noisy data in Center 3) (\%)(30 repeats).}
\label{Tab:SWT_N3}
\begin{tabular}{l|l|c|c|c|c|c|c|c|c|c|c}
\hline
Initialization & Metric & joint & IT & FT & EWC &SI  & MAS & LWF & EBLL & IMM-mean & IMM-mode\\
\hline
\multirow{4}{*}{Random (39.63)} & accuracy& 55.55& 38.86& 47.75 & 48.15 & 47.54 & 42.85 & 46.99 &45.47 & 46.43 & 46.89\\
& standard deviation &1.53 & 1.94 & 2.0 & 1.75 & 1.68 & 1.84& 2.13 & 2.34 & 2.21 & 2.36\\
&monotonicity& n.a.&n.a. & 67.17 & 70.0 & 63.0 & 60.17 & \textbf{74.33} &\textbf{76.67} & \textbf{76.5} & \textbf{76.33}\\
&significance & n.a. & n.a. &n.a.&No & No & Yes- &No & Yes- & No & No\\
\hline
\multirow{4}{*}{Low (34.2)} & accuracy & n.a.& n.a.&  47.43& 47.23 &46.35 & 40.99 & 47.76 & 44.03 &44.97 & 46.47 \\
&standard deviation& n.a. & n.a.&  1.71& 1.84 & 1.46 & 0.97 & 1.41 & 2.11 &1.41 & 1.48 \\
&monotonicity& n.a. & n.a.&  78.67& 75.5& 75.67& 68.5 & \textbf{84.5} & \textbf{80.33} &\textbf{86.33} & \textbf{84.67} \\
&significance & n.a. & n.a. &n.a.&No & No & Yes- &No & Yes- & Yes- & No\\
 \hline
 \multirow{4}{*}{High (44.8)}&accuracy &n.a. & n.a. & 48.0 & 48.31 & 48.11 & 44.9 & 48.49 & 46.56 & 46.6 & 45.04\\
& standard deviation &n.a. & n.a.&1.86 & 1.89 &1.65 & 1.98 & 1.65 & 2.27 &3.74 & 4.38 \\
&monotonicity&n.a. & n.a.& 60.5 & 61.33 & 60.33 & 57.5 & \textbf{72.67} &\textbf{67.67} & \textbf{69.33} & \textbf{68.5}\\
&significance & n.a. & n.a. &n.a.&No & No & Yes- &No & No & No & Yes-\\
 \hline
\end{tabular}
\end{table*}

\begin{table*}
\centering
\caption{The average performances of different methods with CWT for Non-IID Tiny ImageNet data with random 10-epoch initializations (\%)(30 repeats).}
\label{Tab:CWT_N_all}
\begin{tabular}{l|l|c|c|c|c|c|c|c|c|c|c}
\hline
Noisy center & Metric & joint & IT & FT & EWC &SI  & MAS & LWF & EBLL & IMM-mean & IMM-mode\\
\hline
\multirow{4}{*}{Center 5} & accuracy & 55.62 & 39.12 &50.20 & 50.73 & 49.54 & 40.05 & \textbf{54.67} & \textbf{54.90} &  \textbf{54.17} &  \textbf{54.42}\\
& standard deviation &1.42& 1.94 & 1.62 & 1.75 & 1.77 & 3.58 &2.07 & 1.79 &1.79 & 1.80\\
&monotonicity& n.a. & n.a.&  59.80& 58.94 & 59.92& 59.97 & \textbf{67.83} &\textbf{70.08} & \textbf{73.33} & \textbf{71.50}\\
&significance & n.a. & n.a. &n.a.&No & No & Yes- &\textbf{Yes+} & \textbf{Yes+} & \textbf{Yes+} & \textbf{Yes+}\\
\hline
\multirow{4}{*}{Center 3} & accuracy & 55.55 & 38.86 &52.51 & 52.17 & 51.15 & 42.23 & 53.61 & 53.59 &  52.72 &  52.67\\
& standard deviation &1.53& 1.94 & 1.77 & 1.98 & 1.77 & 4.14 &2.10 & 1.85 &1.87 & 1.83\\
&monotonicity& n.a. & n.a.& 60.94& 60.36 & 59.75& 60.17 & \textbf{67.42} &\textbf{67.58} & \textbf{72.61} & \textbf{71.22}\\
&significance & n.a. & n.a. &n.a.&No & Yes- & Yes- &No & No & No & No\\
\hline
\multirow{4}{*}{Center 4 and 5} & accuracy & 54.82 & 37.79 &50.51 & 50.19 & 49.58 & 39.65 & \textbf{52.63} & \textbf{52.78} &  51.67 &  51.25\\
& standard deviation &1.58& 1.90 & 1.98 & 1.83 & 1.52 & 3.34 &2.13 & 1.85 &1.87 & 1.83\\
&monotonicity& n.a. & n.a.& 59.61& 59.14 & 59.0& 58.53 & \textbf{65.39} &\textbf{67.61} & \textbf{73.19} & \textbf{71.06}\\
&significance & n.a. & n.a. &n.a.&No & No & Yes- &\textbf{Yes+} & \textbf{Yes+} & No & No\\
\hline
\end{tabular}
\end{table*}

\subsection{Results of non-IID data}

\subsubsection{\textbf{LWF, EBLL and IMM improve both accuracy and monotonicity for non-IID data}}
The SWT results for non-IID data with noisy data in Center 5 are displayed in Tab.\,\ref{Tab:SWT_N5}. For SWT, the initial model performance matters as described above. When the performances of initial models are randomized, LWF, IMM-mean and IMM-mode achieve slightly better average accuracies than FT. However, when a high-accuracy initial model is used for all the 30 repeats, LWF, EBLL, IMM-mean and IMM-mode all achieve significantly better accuracies than FT, in addition to better monotonicity. The plots of representative accuracy curves for different methods with the high-accuracy initialization are displayed in Fig.\,\ref{Fig:SWTN5}. The curves of FT in Centers 1-4 have obvious accuracy drops after training in Center 5 because of data heterogeneity in Center 5. In contrast, the curves of LWF, EBLL, IMM-mean and IMM-mode in Centers 1-4 have lower magnitudes of accuracy drops after training in Center 5.

The benefit in accuracy improvement of LWF, EBLL, IMM-mean and IMM-mode is more evident in the CWT results for non-IID data (noisy data in Center 5), as displayed in Tab.\,\ref{Tab:CWT_N_all}. The average accuracy of FT is 50.20\%, which is 5.42\% lower than that of the joint method. In contrast, the average accuracies of LWF, EBLL, IMM-mean and IMM-mode are all very close to that of joint training, with a gap of around 1\%. The representative accuracy curves of FT, LWF, EBLL and IMM are displayed in Fig.\,\ref{Fig:CWTN5}, where both FT and EWC have large and frequent oscillations, while LWF, EBLL, IMM-mean and IMM-mode have much lower oscillation magnitudes.

When the noisy data is located in Center 5, the final model tends to overfit to such noisy data and hence has decreased performance on the test data from other centers. This can be observed clearly from Fig.\,\ref{Fig:SWTN5} and Fig.\,\ref{Fig:CWTN5}, where the accuracy of FT at each center typically drops after the model is trained in Center 5. In contrast, LWF and EBLL use teacher-student learning to avoid large deviation from the previous models, while IMM-mean and IMM-mode use model merging to avoid such drastic performance drops. All these methods (LWF, EBLL, IMM-mean and IMM-mode) are able to improve the accuracy smoothly in general after training in Center 5. 

\subsubsection{\textbf{Importance-based methods (EWC, SI and MAS) have no significant benefits}}
The network parameter importance based methods EWC, SI and MAS could not get significantly better performance than FT in our experiments for both IID data (Tab.\,\ref{Tab:SWT_IID} and Tab.\,\ref{Tab:CWT_IID}) and non-IID data (Tab.\,\ref{Tab:SWT_N5} and Tab.\,\ref{Tab:CWT_N_all}). These methods have the plasticity-stability dilemma \cite{grossberg2012studies}: stability is gained with a larger $\lambda$ value in Eqn.\,(\ref{eq:SIcontinualLoss}), while plasticity is gained with a lower $\lambda$ value. In this work, different values for $\lambda$ are checked. For example, for the SWT performance of SI on non-IID data (noisy data in Center 5), SI achieves average accuracies of 48.0\%, 46.97\%, 46.08\% and 44.72\% for $\lambda = 0.1, 1, 10$ and 100, respectively. With a larger $\lambda$ value, a lower average accuracy is obtained but with higher stability, and SI with a low $\lambda$ value of 0.1 has similar performance to FT. For multi-task continual learning scenarios, such importance-based methods constrain important parameter weights in the feature extractor with less change to avoid catastrophic forgetting. For single-task continual learning scenarios in this work, an alternative definition that allows more freedom for important parameter weights but constrains unimportant ones is also investigated with the following modified objective function,
\[
\mathcal{L}\muindex = \mathcal{L}_{\textrm{task}} + \lambda\muindex \sum_k \frac{1}{1 + \boldsymbol{\Omega}\muindexpre_k} \left(\boldsymbol{\theta}\muindexpre_{k} - \boldsymbol{\theta}_k\right)^2.
\label{eq:SIcontinualLoss2}
\]
However, with the modification, EWC, SI and MAS still do not outperform FT. All in all, our experiments demonstrate that importance-based methods (EWC, SI and MAS) have no significant benefits for the single-task multicenter collaboration scenarios in our ITL framework.

\subsubsection{\textbf{The order of centers matters for training with non-IID data}}
When the noisy data is located in Center 5, a significant benefit in accuracy improvement for LWF, EBLL and IMM is observed. However, when the noisy data is located in Center 3, all the continual learning regularization methods are not better than FT, with some (e.g., IMM-mode for SWT with a high-accuracy initialization) even being significantly worse than FT, as displayed in Tab.\,\ref{Tab:SWT_N3} and Tab.\,\ref{Tab:CWT_N_all}. It is worth noting that the accuracies of all the methods with noisy data in Center 3 are lower than those obtained by LWF, EBLL and IMM when the noisy data is in Center 5. This indicates that the order of centers matters for training with non-IID data. When two centers (Center 4 and Center 5) have noisy data, LWF and EBLL can still achieve significantly better accuracies than FT, as demonstrated by the last row of Tab.\,\ref{Tab:CWT_N_all}. Although the 95\% confidence intervals of IMM-mean and IMM-mode overlap with that of FT, their average accuracies are higher than that of FT. It is suggested by our experiments that superior accuracy can be obtained by certain continual learning regularization (LWF, EBLL and IMM) when the centers with heterogeneous data are placed at the end of the training sequence; otherwise no improvement from continual learning regularization is achieved. The fundamental reason is similar to the reliance on initialization: if the center with heterogeneous data is located in the beginning or middle of the training sequence, a suboptimal/inferior model for all the test data is obtained at that center, which influences all the subsequent trainings.

\subsubsection{\textbf{The heterogeneous data is still beneficial to improve performance}}
When Center 5 contains noisy data, training in Center 5 causes accuracy drops for FT. Therefore, it is interesting to know whether training on four centers without Center 5 will lead to superior performance for the overall test in all the five centers. The last row of Tab.\,\ref{Tab:SWT_N5} indicates that FT achieves slightly lower accuracy (four centers: 46.89\% vs. five centers: 47.11\% ) without significance. The same is observed for IMM-mode (48.11\% vs. 48.24\%). However, LWF (47.58\% vs. 49.91\%), EBLL (47.26\% vs. 48.55\%) and IMM-mean (49.25\% vs 50.67\%) all obtain \textbf{significantly} lower accuracies when only data from four centers is used for training. Therefore, the heterogeneous data in Center 5 is still beneficial to improve performance for LWF, EBLL and IMM-mean.
Note that the significance here is computed between four centers and 5 centers, while the significance in Tab.\,\ref{Tab:SWT_N5} is computed between each method and FT.

\begin{figure}
\centering
\includegraphics[width=\linewidth]{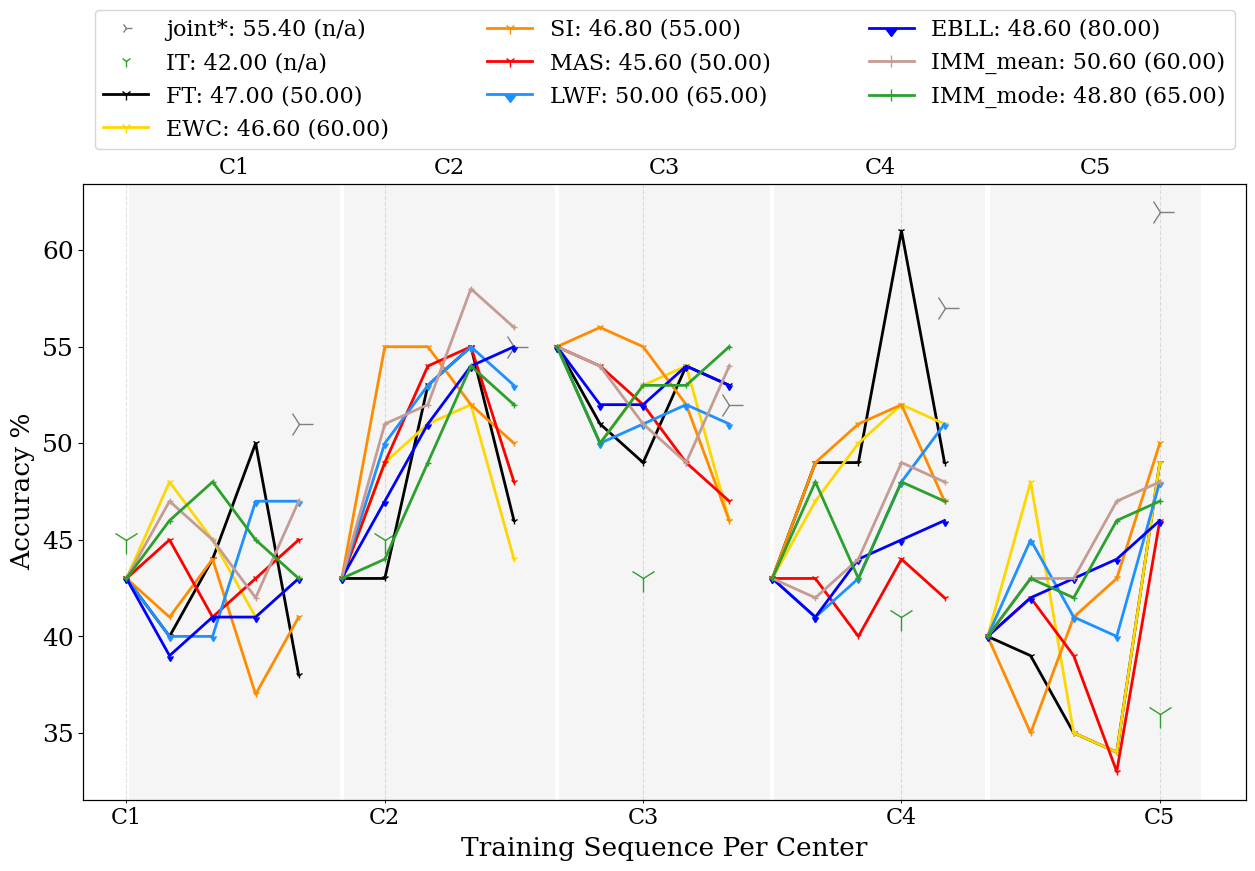}
\caption{The SWT results of continual learning regularization methods in ITL on Tiny ImageNet data, where the data in Center 5 contains noise. }
\label{Fig:SWTN5}
\end{figure}

\begin{figure*}
\centering
\includegraphics[width=1.0\linewidth]{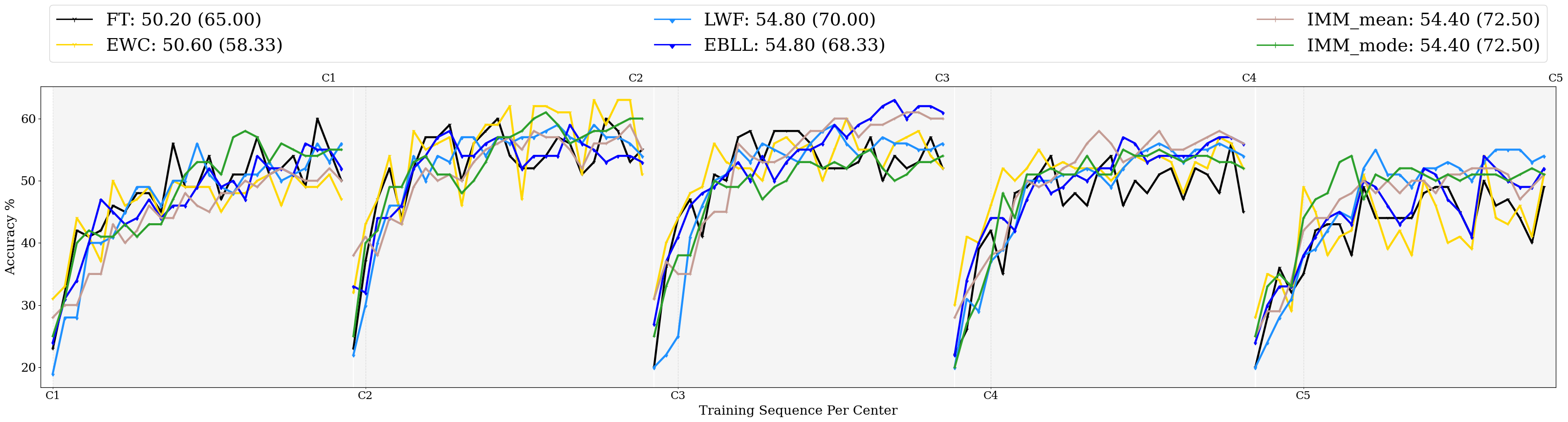}
\caption{The representative CWT results of continual learning regularization methods on the Tiny ImageNet data (with different initial models), where the data in Center 5 contains noise. LWF, EBLL, IMM-mean, and IMM-mode demonstrate a lower degree of oscillations than FT and EWC. The joint model has an average accuracy of 55.62\%. The curves of SI and MAS are similar to that of EWC and hence are omitted for better visualization.}
\label{Fig:CWTN5}
\end{figure*}

\section{Experiments on retinal fundus data for diabetic retinopathy classification}
To demonstrate the efficacy of the continual learning methods on real medical data, all the methods are applied to retinal fundus images for diabetic retinopathy classification and T1 contrast enhanced MRI images for glioma segmentation (Section \ref{Sect:Glioma}). In this diabetic retinopathy classification experiment, the data is split based on imaging devices where data heterogeneity is clearly observed, which is a realistic data split for multicenter collaboration.

\subsection{Experimental Setup}
\subsubsection{Dataset}
\begin{figure}
\centering
\begin{minipage}{0.3\linewidth}
\subfigure[$1424 \times 2144$, 0]{
\includegraphics[width=\linewidth]{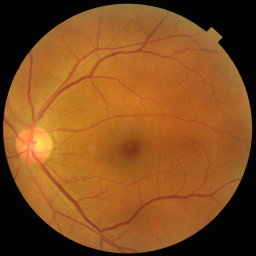}
}
\end{minipage}
\begin{minipage}{0.3\linewidth}
\subfigure[$1424 \times 2144$, 1]{
\includegraphics[width=\linewidth]{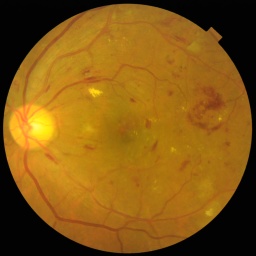}
}
\end{minipage}
\begin{minipage}{0.3\linewidth}
\subfigure[$2848 \times 4288$, 1]{
\includegraphics[width=\linewidth]{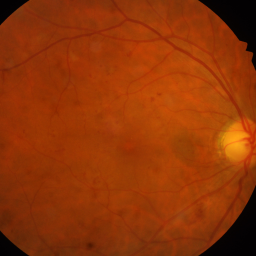}
}
\end{minipage}

\begin{minipage}{0.3\linewidth}
\subfigure[$2000 \times 3008$, 1]{
\includegraphics[width=\linewidth]{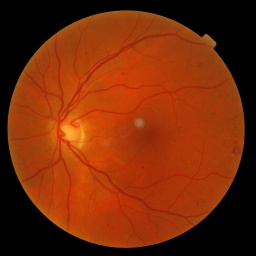}
}
\end{minipage}
\begin{minipage}{0.3\linewidth}
\subfigure[$3264 \times 4928$, 1]{
\includegraphics[width=\linewidth]{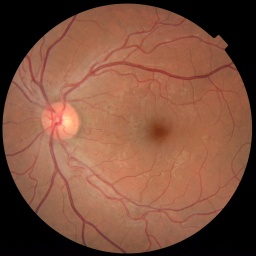}
}
\end{minipage}
\begin{minipage}{0.3\linewidth}
\subfigure[$3264 \times 4928$, 0]{
\includegraphics[width=\linewidth]{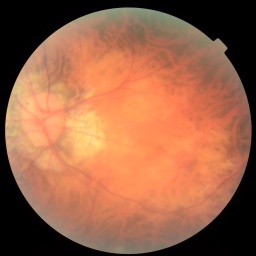}
}
\end{minipage}
\caption{Exemplar retinal fundus images from different cameras (same size from the same camera). The original image sizes before preprocessing are displayed in the subcaptions, followed by the corresponding diabetic retinopathy classification (0: negative; 1: positive). In (f), other unclassified retinal disease exists, although it is diabetic retinopathy negative.}
\label{Fig:RetinaExamples}
\end{figure}

The retinal fundus images used in this work are from two sources: the retinal fundus multi-disease image dataset (RFMiD) \cite{pachade2021retinal} and the Kaggle diabetic retinopathy detection dataset (KDRDD)\footnote{\url{https://www.kaggle.com/competitions/diabetic-retinopathy-detection/data}}. Such images are acquired from different imaging systems with different cameras and hence have different sizes. All the images are cropped and resized to $256 \times 256$ images. The images are further divided into different centers according to their original image sizes. In such a way, images at different centers have different styles, which reflects data heterogeneity in a realistic manner. In this work, a binary classification of diabetic retinopathy is evaluated. Five centers are chosen for the experiments in this work and the details of the data at each center are displayed in Tab.\,\ref{Tab:Retina}. Each center has a data split of 64\%, 16\%, and 20\% for training, validation, and test. Some exemplary images are displayed in Fig.\,\ref{Fig:RetinaExamples} along with their original image size and diabetic retinopathy classification. Note that for images from the source of KDRDD, diabetic retinopathy negative images may contain other unclassified retinal diseases like Fig.\,\ref{Fig:RetinaExamples}(f). Therefore, the images from KDRDD are noisier than those from RFMiD. The symptoms of diabetic retinopathy include small bulges in blood vessels, blood leakages, deposits of cholesterol, irregular beading, obliteration of blood vessels and growth of new blood vessels \cite{asia2022detection}. For mild cases, the symptoms are difficult to observe for human being (e.g., Figs.\,\ref{Fig:RetinaExamples}(d) and (e)) and hence the automatic classification of diabetic retinopathy has important clinical value.

\begin{table*}
\caption{The details of retinal fundus images in the five centers.}
\label{Tab:Retina}
\centering
\begin{tabular}{lccccc}
\hline
Center & Center 1 & Center 2 & Center 3 & Center 4 & Center 5 \\
\hline
Data source & RFMiD & RFMiD & KDRDD & KDRDD & KDRDD\\
\hline
Original size & $1424 \times 2144$ & $2848 \times 4288$ & $2848 \times 4288$ & $2000 \times 3008$ & $3264 \times 4928$ \\
\hline
Total number & 852 & 254 & 155 & 398 & 720\\
\hline
Number of negative & 333 & 142 & 122 & 326 & 512\\
\hline
Number of positive & 519 & 112 & 33 & 72 & 208\\
\hline
\end{tabular}
\end{table*}

\begin{table*}
\centering
\caption{The average performances of different methods with SWT for retinal data (\%)(30 repeats).}
\label{Tab:SWT_retina}
\begin{tabular}{l|l|c|c|c|c|c|c|c|c|c|c}
\hline
Initialization & Metric & joint & IT & FT & EWC &SI  & MAS & LWF & EBLL & IMM-mean & IMM-mode\\
\hline
\multirow{4}{*}{Random (53.12)} & accuracy& 79.69 & 44.69 (64.86)& 57.21 & 54.57& 56.87 & 55.30 & \textbf{61.09} &\textbf{59.85} & \textbf{61.22} & \textbf{63.76}\\
& standard deviation &4.28 & 3.25 (5.45) & 10.02 & 8.23 & 9.46 & 9.79 & 8.42 & 7.33 & 9.86 & 10.21\\
&monotonicity& n.a.&n.a. & 55.50 & 53.0 & 53.67 & 53.17 & \textbf{63.83} &\textbf{67.83} & \textbf{63.0} & \textbf{63.67}\\
&significance & n.a. & n.a. &n.a.&No & No & No &No & No & No & No\\
 \hline
 \multirow{4}{*}{High (65.29)}&accuracy &n.a. & n.a. & 61.37 & 63.12 & 63.78 & \textbf{66.06}& \textbf{71.11} & \textbf{69.58} & \textbf{68.83}& \textbf{70.27}\\
& standard deviation &n.a. & n.a.&5.71 & 5.34 & 5.22 & 3.45 & 4.23 &3.13 & 2.67 & 2.38\\
&monotonicity&n.a. & n.a.& 47.50 & 47.33 & 48.17 & 49.83 & \textbf{65.50} &\textbf{64.67} & \textbf{58.33} & \textbf{58.50}\\
&significance & n.a. & n.a. &n.a.&No & No & \textbf{Yes+} &\textbf{Yes+} & \textbf{Yes+} & \textbf{Yes+} & \textbf{Yes+}\\
 \hline
\end{tabular}
\end{table*}

\begin{table*}
\centering
\caption{The average performances of different methods with CWT for retinal data (\%)(30 repeats).}
\label{Tab:CWT_Retina}
\begin{tabular}{l|l|c|c|c|c|c|c|c|c|c|c}
\hline
Initialization & Metric & joint & IT & FT & EWC &SI  & MAS & LWF & EBLL & IMM-mean & IMM-mode\\
\hline
\multirow{2}{*}{Random (54.25)} & accuracy & 79.69& 44.69 (64.86) &72.45 & 73.26 & 70.69 & 71.90 & \textbf{74.35} & \textbf{74.22} &  \textbf{74.53} &  \textbf{74.96}\\
& standard deviation &4.28& 3.25 (5.45) & 2.53& 3.61 & 2.89 & 3.10 & 2.36 &2.82 & 3.02 & 2.53\\
\multirow{2}{*}{(15 epochs)}& monotonicity & n.a. & n.a.&  52.17& 52.98 & 50.48& 58.26 & \textbf{63.61} &\textbf{83.42} & \textbf{61.42} & \textbf{61.08}\\
&significance & n.a. & n.a. &n.a.&No & No & No &\textbf{Yes+} & \textbf{Yes+} & \textbf{Yes+} & \textbf{Yes+}\\
\hline
\end{tabular}
\end{table*}

\subsubsection{Network}
The same small VGG-9 network for the Tiny ImageNet data is used for diabetic retinopathy classification. For SWT, all the models are trained for 100 epochs at each center. For monitoring overfitting, the learning rate is decayed into half after every $E_\text{val} = 5$ epochs and early stopping is executed after $E_\text{stop} = 20$ epochs.  For CWT, a weight transfer frequency of $E_\text{transfer}=15$ is investigated with a total number of 5 iterations. The batch size is 10. The Adam optimizer is used for training with an initial learning rate of 0.001. Other training parameters are the same as those for the Tiny ImageNet data. It is worth noting that the weighted random sampler from Pytorch is applied to get a balanced class distribution for each training and validation batch, otherwise it is difficult for the network to learn due to class imbalance.

\subsection{Results}
The SWT results with different algorithms are displayed in Tab.\,\ref{Tab:SWT_retina}. The joint model with mixed training data achieves an average accuracy of 79.69\% among the 30 repeats. IT achieves an average accuracy of 64.86\% (the model in each center is trained and tested with its own data), which is relatively high. This is because each local model is optimized for the specific data from the same center. But such a model has very poor generalizability to test data in other centers, e.g., the local model from Center 5 achieves an average accuracy of 44.69\% for test datasets from all centers. With all random initializations, FT achieves an average accuracy of 57.21\%. LWF, EBLL, IMM-mean and IMM-mode have achieved higher average accuracies than FT, but without statistical significance. Picking a fixed high-accuracy (65.29\%) model for the first center and continuing stochastic training subsequently for 30 repeats, LWF, EBLL, IMM-mean and IMM-mode have significantly better accuracies than FT, which is consistent with the observation from the Tiny ImageNet experiments.

The CWT results are displayed in Tab.\,\ref{Tab:CWT_Retina}. With CWT, FT achieves a better accuracy of 72.45\%, which is 15.24\% higher than that with SWT. The importance weight constrained methods (EWC, SI, and MAS) again have comparable accuracies to FT. Consistently, LWF, EBLL, IMM-mean and IMM-mode have higher accuracies than FT, among which LWF, IMM-mean and IMM-mode have statistical significance ($p < 0.05$).

\section{Experiments on glioma segmentation}
\label{Sect:Glioma}
To demonstrate the efficacy of continual learning methods on real medical image segmentation tasks (in addition to the image classification tasks), the methods are also investigated on the glioma segmentation using multicenter data.

\subsection{Experimental Setup}
\subsubsection{Dataset}
\begin{figure}
\centering

\begin{minipage}{0.45\linewidth}
\subfigure[UPenn-GBM, gradient echo]{
\includegraphics[width=\linewidth]{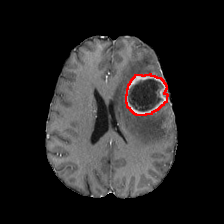}
\label{subfig:UPenn}
}
\end{minipage}
\begin{minipage}{0.45\linewidth}
\subfigure[UCSF-PDGM, gradient echo]{
\includegraphics[width=\linewidth]{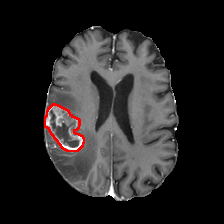}
\label{subfig:UCSF_MPRAGE}
}
\end{minipage}

\begin{minipage}{0.45\linewidth}
\subfigure[BraTS 2020, gradient echo]{
\label{subfig:BraTS_MPRAGE}
\includegraphics[width=\linewidth]{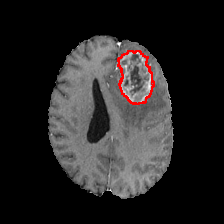}
}
\end{minipage}
\begin{minipage}{0.45\linewidth}
\subfigure[BraTS 2020, spin echo]{
\includegraphics[width=\linewidth]{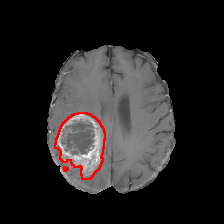}
\label{subfig:BraTS_SE}
}
\end{minipage}
\caption{Exemplar T1 contrast enhanced MRI images from different centers in our experiments. Center 1 to 3 ((a) to (c)) use gradient echo MRI sequences, while Center 4 uses a spin echo MRI sequence. The  UPenn-GBM (a) and BraTS 2020 (c) images are noisier than the UCSF-PDGM (a) images. The red curves contour the reference tumor cores.}
\label{Fig:GliomaExamples}
\end{figure}

The glioma datasets are from three sources: BraTS 2020 data \cite{menze2014multimodal}, the UCSF-PDGM dataset \cite{calabrese2022university}, and the UPenn-GBM dataset \cite{bakas2022university}. The BraTS 2020 dataset contains multicentric data, but such center information for each patient's data is not available to users. Therefore, in our study, the BraTS 2020 dataset is simply split into two virtual centers based on the two basic T1 contrast enhanced MRI sequences gradient echo (e.g., MPRAGE) and spin echo (e.g., SPACE). In spin-echo images, the contrast between gray/white matters and the contrast for blood vessels are less than those in gradient echo images, as displayed in Fig.\,\ref{subfig:BraTS_MPRAGE} and Fig.\,\ref{subfig:BraTS_SE}, respectively. The images from UCSF-PDGM and UPenn-GBM both use the gradient echo T1 contrast sequences. However, the UCSF-PDGM dataset is less noisy than the UPenn-GBM dataset and the BraTS 2020 dataset, as displayed in Fig.\,\ref{Fig:GliomaExamples}. All the volumes have $240 \times 240 \times 155$ voxels with an isotropic spacing of 1\,mm. The BraTS glioma segmentation challenge aims to segment normal background tissues, necrotic tumor core, contrast-enhancing tumor, and peritumoral edema. The tumor core consists of necrotic tumor core and contrast-enhancing tumor. As tumor core frequently represents the only relevant tumor compartment for clinical treatment planning \cite{putz2023segment,bernhardt2022degro,wittenstein2023tumor}, in this work we focus on tumor core segmentation on contrast-enhanced T1-weighted images. To save computation, the volumes are cropped to $224 \times 224$ 2D images with an isotropic spacing of 1\,mm and one slice among 10 neighboring slices is selected to avoid similarity. 
For the four centers, the numbers of training images are 150, 262, 206, and  122, respectively; The numbers of validation images are 54, 122, 120, and 122, respectively; The numbers of test images are 170, 500, 160, and 158, respectively.

\subsubsection{Network}
The U-Net from \cite{buda2019association} is used for glioma segmentation. It consists of 5 levels. The Dice coefficient loss function is used for training and validation. For SWT, all the models are trained for 100 epochs at each center. For monitoring overfitting, the learning rate is decayed into half after every $E_\text{val} = 5$ epochs and early stopping is executed after $E_\text{stop} = 20$ epochs. For CWT, a weight transfer frequency of $E_\text{transfer}=15$ is investigated with a total number of 5 iterations. The batch size is 16. The Adam optimizer is used for training with an initial learning rate of 0.001. A constant $\lambda$ value of 0.1 instead of 1 for LwF, EWC, SI and MAS is used. Due to the similar performance of EBLL to LWF but with much more computation for training each encoder at each iteration, for the glioma segmentation EBLL is omitted. Other training parameters are the same as those for the Tiny ImageNet data.

\begin{table*}
\centering
\caption{The average performances of different methods with SWT and CWT for glioma data (\%)(30 repeats).}
\label{Tab:SWT_glioma}
\begin{tabular}{l|l|c|c|c|c|c|c|c|c|c}
\hline
Weight transfer & Metric & joint & IT & FT & EWC &SI  & MAS & LWF  & IMM-mean & IMM-mode\\
\hline
\multirow{4}{*}{SWT} & Dice coefficient& 73.81 &  59.03 (68.31)& 65.56 & 64.23& 65.02 & 64.44 & \textbf{66.44}  & \textbf{67.07} & \textbf{66.38}\\
& standard deviation &1.33 &  1.26 (1.01) & 1.53 & 1.50 & 1.85 & 2.71 & 1.36 & 3.66 & 4.87\\
&monotonicity& n.a.&n.a. & 53.12 & 53.0 & 53.67 & 53.17 & 55.24 & \textbf{69.27} & \textbf{70.83}\\
&significance & n.a. & n.a. &n.a.&No & Yes- & Yes- &\textbf{Yes+} & No & No\\
 \hline
 \multirow{4}{*}{CWT}&Dice coefficient &n.a. & n.a. & 67.08 & 66.56 & 67.06 & 66.76& \textbf{73.45} & 68.06 & 67.85\\
& standard deviation &n.a. & n.a.&2.04 & 2.46 & 2.72 & 3.42 & 1.19 & 4.52 & 4.36\\
&monotonicity&n.a. & n.a.& 54.47 & 55.06 & 55.89 & 52.38& \textbf{64.96}  & 52.15 & 53.50\\
&significance & n.a. & n.a. &n.a.&No & No & No &\textbf{Yes+} & No & No\\
 \hline
\end{tabular}
\end{table*}

\subsection{Results}
\begin{figure}
\centering
\includegraphics[width=\linewidth]{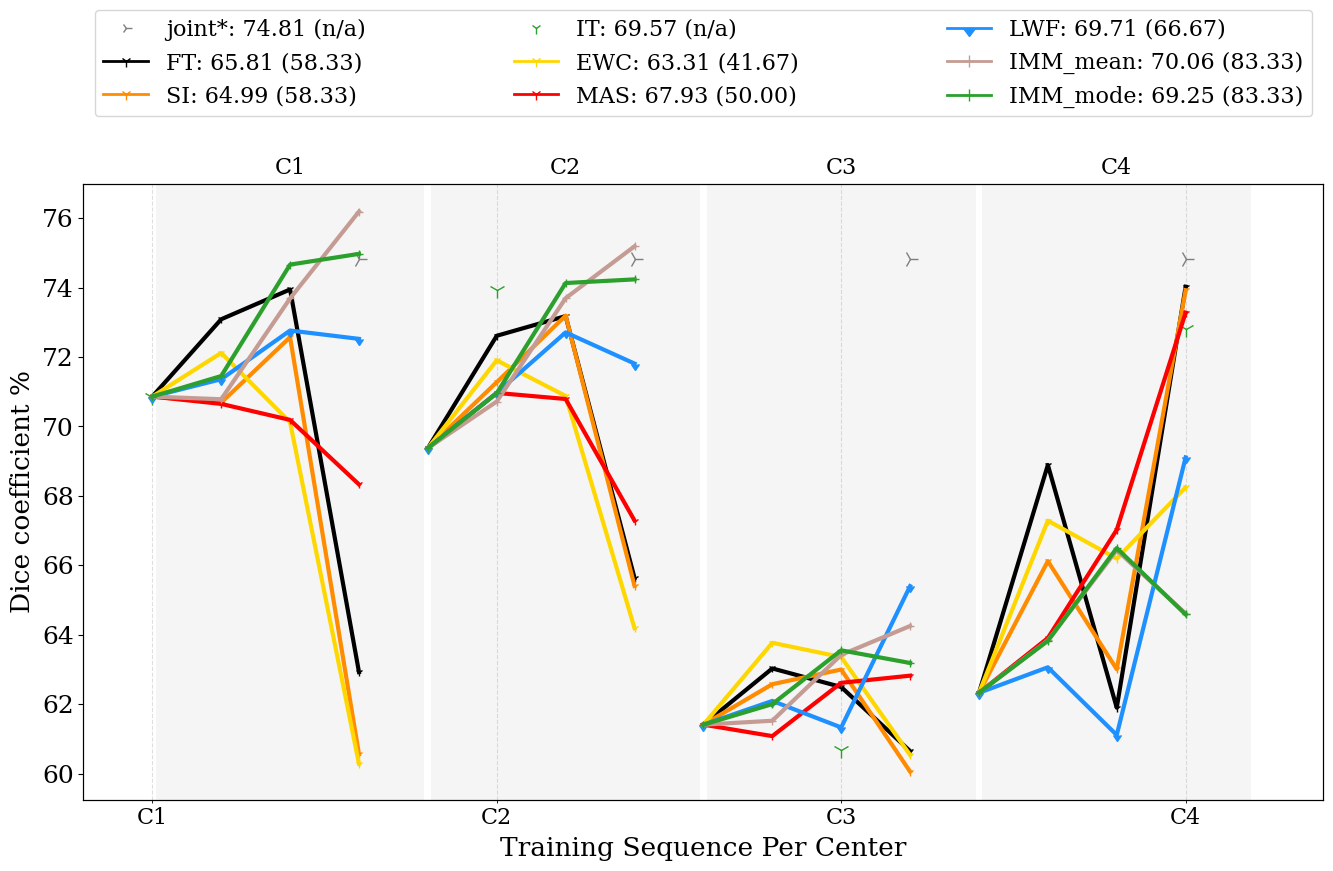}
\caption{The exemplary SWT results of continual learning regularization methods on glioma data. }
\label{Fig:GliomaSWT}
\end{figure}

The SWT results with different algorithms are displayed in Tab.\,\ref{Tab:SWT_glioma}. The joint model with mixed training data achieves an average Dice coefficient of 0.7381 among the 30 repeats. Among all the methods, LWF, IMM-mean and IMM-mode have higher average Dice coefficients than FT, which is consistent with our observation from the Tiny ImageNet data and the retinal data. Among them, LWF is significantly better than FT, while IMM-mean and IMM-mode do not have significance due to their large variations. The exemplary plots of different SWT methods are displayed in Fig.\,\ref{Fig:GliomaSWT}. For the FT curve in Center 1, its Dice coefficient increases after subsequent training in Center 2 and Center 3, but it has a drastic decrease after training in Center 4. In contrast, LWF, IMM-mean and IMM-mode do not have such drastic drops after training in Center 4 (because data in Center 4 uses the T1 spin echo instead of T1 gradient echo sequence). Note that Center 3 (BraTS 2020, gradient echo) has relatively low Dice coefficients for all models because the BraTS 2020 gradient echo dataset contains gliomas of different stages and meanwhile the data is noisy. In contrast, the UPenn-GBM dataset contains glioblastomas which typically have larger sizes than those in the BraTS dataset, and the UCSF-PDGM dataset is less noisy. 

\begin{figure*}
\centering
\includegraphics[width=\linewidth]{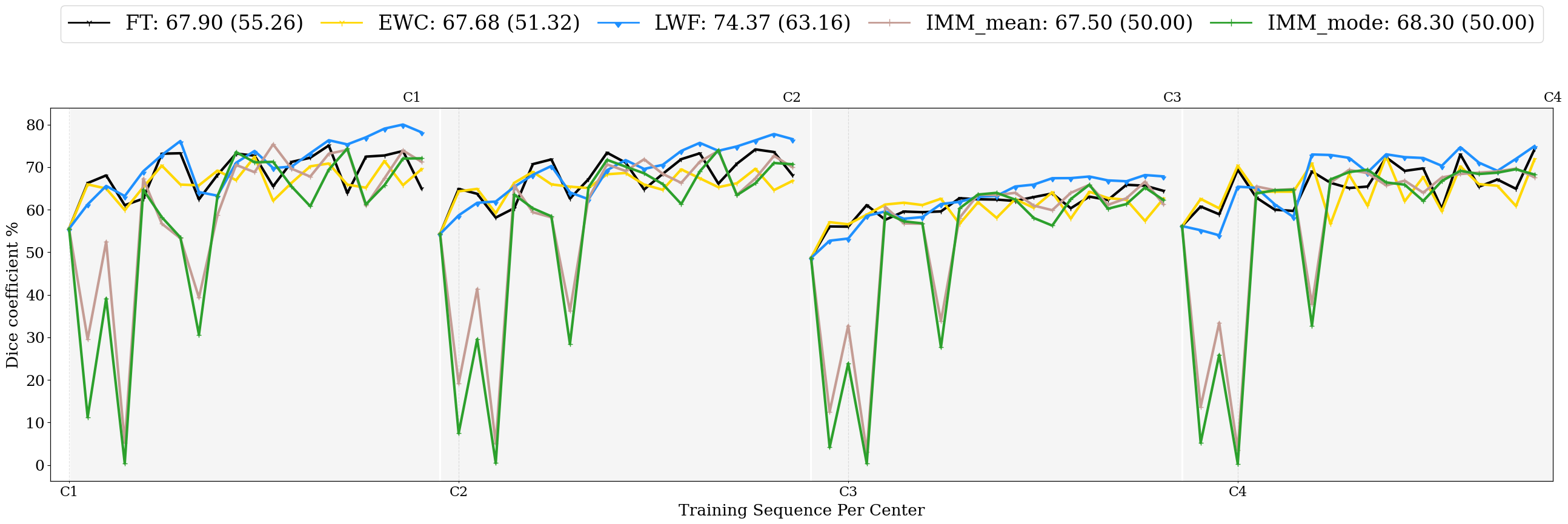}
\caption{The exemplary CWT results of continual learning regularization methods on glioma data. LWF demonstrates a lower degree of oscillations than FT. IMM-mean and IMM-mode have severe oscillations for in the early transfers. The curves of SI and MAS are similar to that of EWC and hence are omitted for better visualization.}
\label{Fig:GliomaCWT}
\end{figure*}

The CWT results are displayed in Tab.\,\ref{Tab:SWT_glioma} as well. LWF has a significantly better average Dice coefficient 0.7345 in comparison to that of FT 0.6708. Although IMM-mean and IMM-mode have better average Dice coefficients, their values are very close to that of FT. The importance weight constrained methods (EWC, SI, and MAS) again have comparable Dice coefficients to FT. The exemplary plots of different CWT methods are displayed in Fig.\,\ref{Fig:GliomaCWT}. EWC has reduced oscillations compared with FT. However, its final Dice coefficient is not better than FT. LWF has reduced oscillations with stably improved Dice coefficients. It is interesting to observe that IMM-mean and IMM-mode have very large performance drops at the early transfers, while it recovers to a relatively stable state in the late transfers. For segmentation tasks, the model needs a pixel-level classification instead of an image-level classification. Therefore, the final output of a segmentation model is much more sensitive to the model weights than that of a classification model. For IMM-mean and IMM-mode with CWT, these methods merge model weights from previous centers and hence there is a risk that the merged models have a low performance. Since each model reaches a relatively better local minimum in SWT than that in CWT in the early transfers, such performance drop in SWT is less frequently observed than that in CWT in our experiments.

\section{Discussion}

\subsection{Data heterogeneity and center order}
Our experiments demonstrate that with our proposed ITL framework, a certain level of forgetting occurs for FT in the IID-data case, but this forgetting is not catastrophic as that in task incremental learning. However, with non-IID data, catastrophic forgetting occurs when the model is trained in the centers with  heterogeneous data. In practice, whether the datasets from multiple centers are heterogeneous or not can be determined by prior information before training (e.g., the imaging device and preprocessing pipeline information) or be computed in feature space (e.g. using t-distributed stochastic neighbor embedding (t-SNE) \cite{hinton2002stochastic} to visualize feature distributions of datasets from different centers, or training a small encoder like EBLL does). In addition to this, some SWT or CWT trials using FT can be performed to check the data heterogeneity. Based on the FT results (e.g., Fig.\,\ref{Fig:SWTN5}, Fig.\,\ref{Fig:CWTN5}, Fig.\,\ref{Fig:GliomaSWT} and Fig.\,\ref{Fig:GliomaCWT}), if the model performance drops frequently at certain centers, the datasets from these centers are likely to have different features or styles from the majority of centers. Therefore, they should be placed at the end of the center sequence for training.

\subsection{Robustness to data privacy attacks}
In federated learning, although only model parameters or training gradients are shared, there is still risk of data leakage. Recent attack techniques, e.g. model inversion attacks, have the capability of reverse engineering private data at federated training phases \cite{huang2021evaluating} and federated inference phases \cite{he2019model,jia2019memguard}. However, such attack methods typically work on C2PFL with split network architectures, where intermediate features of input data are shared among different participants. The layer depth of shared features are one key indicator on how accurate input data can be restored by such model inversion attacks: the shallower the easier \cite{huang2021evaluating}. In addition, such attacks typically are effective for low-resolution images, like the MNIST and CIFER10 datasets. Restoration of high-resolution images from deep networks is much more challenging and requires accurate training statistics, e.g., the batch normalization statistics. In this work, because of the peer-to-peer model sharing strategy, no intermediate features of input (test nor training) images are shared. 
Therefore, restoring private data information is extremely challenging for existing attack algorithms.

\section{Conclusion}
The following conclusions can be drawn from this work:
\begin{itemize}
\item With the multi-head setting, naive weight transfer has the risk of  catastrophic forgetting for both IID data and non-IID data, and all the continual learning regularization methods are effective to avoid catastrophic forgetting.

\item For naive weight transfer, the catastrophic forgetting problem in the multi-head setting can be reduced to a large degree by using the single-head setting.

\item For IID data, because of the reduced forgetting problem by the single-head setting, the continual learning regularization methods cannot obtain significantly better performance than FT (both SWT and CWT).

\item For non-IID data, apparent forgetting pattern is observed after training on centers with heterogeneous data. LWF, EBLL, IMM-mean and IMM-mode are effective to reduce performance oscillations and overall performance drops for both SWT and CWT.

\item For SWT, a high-accuracy initial model trained from the first center is preferred for the guidance of the subsequent training. On the contrary, for CWT a non-convergent (low-accuracy) initial model is preferred.

\item For non-IDD data, the order of centers matters. In particular, the centers with heterogeneous data are preferred to be placed in the last positions for training.

\item For both SWT and CWT with non-IID data, the heterogeneous data is still beneficial to improve the overall performance of the shared model.

\item CWT is better than SWT in the overall accuracy, at the cost of more frequent weight transfers (communications among centers).
\end{itemize}

\

\appendix
\subsection{Additional CWT Results on Tiny ImageNet Data}
The CWT results on the Tiny Image data with converged models as the initialization are displayed in Tab.\,\ref{Tab:CWT_IID_App} and Tab.\,\ref{Tab:CWT_N5_App} for IID data and non-IID data, respectively. 

\subsection{SI with the Adam optimizer}
\label{subsect:SI_Adam}
Taking Adam as an example, Adam keeps two bias-corrected moments $\vm(t)$ and $\vv(t)$ to compute adaptive learning rates for different parameters,
\begin{equation}
\begin{array}{l}
\vm(t)=\left(\beta_1 \vm(t-1) + (1-\beta_1)\vg(t)\right)/\left(1 - \beta_1^t\right),\\
\vv(t) = \left(\beta_2 \vv(t-1) + (1-\beta_2)\vg^2(t)\right)/\left(1 - \beta_2^t\right),
\end{array}
\label{Eqn:Adam}
\end{equation}
 where $\boldsymbol{g}(t) = \partial \boldsymbol{L_{\textrm{task}}}/{\partial \boldsymbol{\theta}\left(t\right)}$ is the gradient of $L_{\textrm{task}}$ with respect to the real-time parameter $\boldsymbol{\theta}\left(t\right)$, and $\beta_1$ and $\beta_2$ are two parameters for $\vm(t)$ and $\vv(t)$, respectively. The $k$-th parameter $\vtheta_k(t)$ is updated as the following,
\begin{equation}
\vtheta_k(t+1) = \vtheta_k(t) - \frac{r}{\sqrt{\vv_k(t)} + \epsilon}\vm_k(t),
\end{equation}
where $r$ is the fixed learning rate and $\epsilon$ is a small value to avoid division by zero, while $\frac{r}{\sqrt{\vv_k(t)} + \epsilon}$ is an adaptive learning rate automatically adjusted for each parameter. 

When continual learning regularization is included, the gradient $\vg(t)$ in Eqn.\,(\ref{Eqn:Adam}) need to be replaced by the new gradient with regularization $\vg'_k(t)$. Taking SI as an example, $\vg'_k(t)$ is modified from $\vg(t)$ as follows according to Eqn.\,(\ref{eq:SIcontinualLoss}),
\begin{equation}
\vg'_k(t) = \vg_k(t) + 2 \lambda \boldsymbol{\Omega}\muindexpre_k \left(\boldsymbol{\theta}\muindexpre_{k} - \boldsymbol{\theta}_k\right).
\end{equation}

In addition, the contribution of each network parameter to the change of the task loss $L_{\textrm{task}}$ need to be tracked in the variable $\boldsymbol{w}_k\muindex$ in the Adam optimizer,
\begin{equation}
\boldsymbol{w}_k\muindex = \int_{t^{\mu-1}}^{t^\mu}\boldsymbol{g}_k\left(\boldsymbol{\theta}\left(t\right)\right)\cdot \boldsymbol{\theta}'_k\left(t\right)\textrm{d}t,
\end{equation}
where $\boldsymbol{\theta}'\left(t\right) = {\partial \boldsymbol{\theta}}/{\partial t}$ is the network parameter change over time. After training in the current center $\mu$, the parameter importance factor $\boldsymbol{\Omega}\muindex$ is updated  as the following for SI \cite{zenke2017continual},
\begin{equation}
\boldsymbol{\Omega}\muindex_k = \sum_{\nu < \mu} \frac{\boldsymbol{w}_k^\nu}{\left(\boldsymbol{\theta}_k^\nu - \boldsymbol{\theta}_k^{\nu-1}\right)^2 + \epsilon},
\label{Eqn:importanceWeight}
\end{equation}
where $\nu$ is the index of a previous center and $\boldsymbol{\theta}^\nu$ is the parameter set of the final trained model at center $\nu$. Hence, the knowledge learned from previous centers are accumulated in this importance factor.

\begin{table*}
\centering
\caption{The average performances of different methods with CWT for IID Tiny ImageNet data (\%)(30 repeats).}
\label{Tab:CWT_IID_App}
\begin{tabular}{l|l|c|c|c|c|c|c|c|c|c|c}
\hline
Initialization & Metric & joint & IT & FT & EWC &SI  & MAS & LWF & EBLL & IMM-mean & IMM-mode\\
\hline
\multirow{2}{*}{Low (35.0)} & accuracy & n.a.& n.a.& 53.10 & 52.71 & 52.37 & 47.75 & 53.65 & 50.57 & \textbf{54.90} & \textbf{54.52}\\
& standard deviation &n.a. & n.a.&1.71 & 1.72 & 1.52 & 1.85 & 1.40 &2.82 & 1.77 & 1.59\\
\multirow{2}{*}{(50 epochs)}&monotonicity& n.a. & n.a.&  60.92& 60.75 & 60.64& 64.88 & 66.33 &77.47 & 68.94 & 67.36\\
&significance & n.a. & n.a. &n.a.&No & No & Yes- &No & Yes- & \textbf{Yes+} & \textbf{Yes+}\\
 \hline
 \multirow{2}{*}{High (45.2)}&accuracy &n.a. & n.a. & 54.77 & 54.17 & 53.11 & 50.25 & 54.57 & 51.47 & \textbf{56.02} & \textbf{55.20}\\
& standard deviation &n.a. & n.a.&1.45 & 1.38 & 1.53 & 0.81 & 1.16 &2.16 & 1.51 & 1.34\\
\multirow{2}{*}{(50 epochs)}&monotonicity&n.a. & n.a.& 57.64 & 57.83 & 57.27 & 65.50 & 65.23 &75.60 & 65.30 & 63.77\\
&significance & n.a. & n.a. &n.a.&No & Yes- & Yes- &No & Yes- & \textbf{Yes+} & No\\
\hline
\end{tabular}
\end{table*}

\begin{table*}
\centering
\caption{The average performances of different methods with CWT for non-IID Tiny ImageNet data (noisy data in Center 5) (\%)(30 repeats).}
\label{Tab:CWT_N5_App}
\begin{tabular}{l|l|c|c|c|c|c|c|c|c|c|c}
\hline
Initialization & Metric & joint & IT & FT & EWC &SI  & MAS & LWF & EBLL & IMM-mean & IMM-mode\\
\hline
\multirow{2}{*}{Low (34.4)} & accuracy & n.a.& n.a.& 52.11 & 52.24 & 51.81 & 44.65 & 52.49 & \textbf{53.45} & \textbf{53.63} & \textbf{53.36}\\
& standard deviation &n.a. & n.a.&1.65 & 0.99 & 1.52 & 1.92 & 1.74 &1.68 & 1.23 & 1.25\\
\multirow{2}{*}{(50 epochs)}&monotonicity& n.a. & n.a.&  59.08& 60.19 & 60.67& 58.69 & \textbf{69.47} &\textbf{69.39} & \textbf{70.92} & \textbf{68.83}\\
&significance & n.a. & n.a. &n.a.&No & No & Yes- &No & \textbf{Yes+} & \textbf{Yes+} & \textbf{Yes+}\\
 \hline
 \multirow{2}{*}{High (43.8)}&accuracy &n.a. & n.a. & 50.75 & 50.95 & 50.01 & 48.83 & \textbf{52.50} & \textbf{52.3} & \textbf{52.99} & \textbf{52.67}\\
& standard deviation &n.a. & n.a.&1.23 & 0.90 & 1.19 & 0.56 & 0.95 &1.00 & 1.03 & 0.87\\
\multirow{2}{*}{(50 epochs)}&monotonicity&n.a. & n.a.& 55.97 & 55.89 & 56.08 & 60.03 & \textbf{64.78} &\textbf{65.14} & \textbf{65.50} & \textbf{65.67}\\
&significance & n.a. & n.a. &n.a.&No & No & Yes- &\textbf{Yes+} & \textbf{Yes+} & \textbf{Yes+} & \textbf{Yes+}\\
\hline
\end{tabular}
\end{table*}

\begin{IEEEbiography}[{\includegraphics[width=1in,height=1.25in,clip,keepaspectratio]{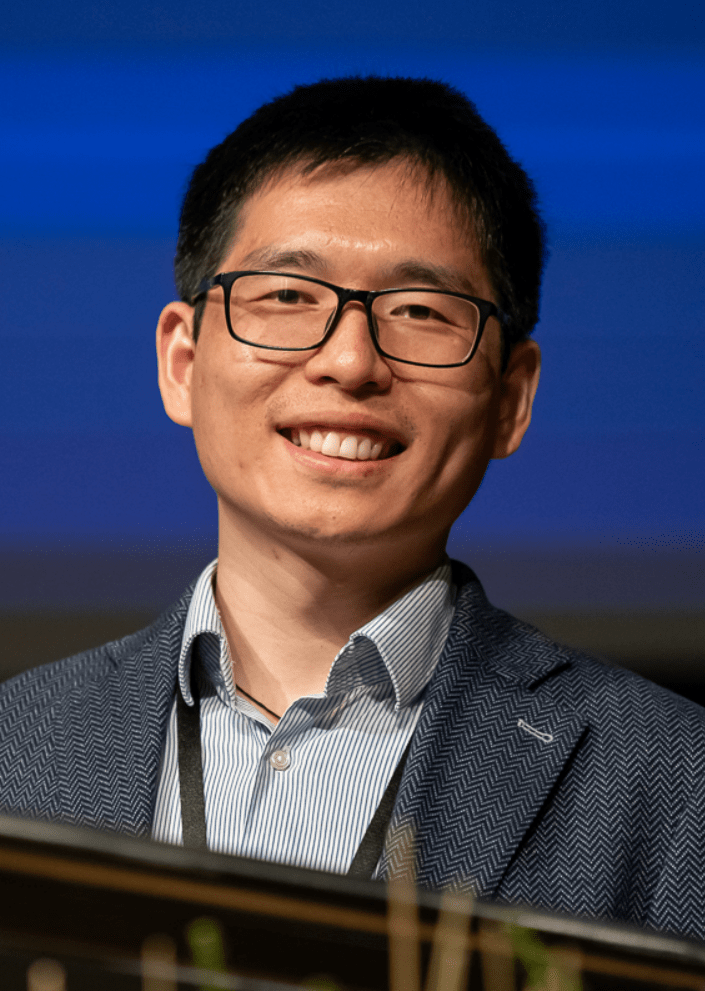}}]{Yixing Huang}
is currently a leading computer science researcher with Department of Radiation Oncology, University Hospital Erlangen, Friedrich-Alexander Universit\"at Erlangen-N\"urnberg (FAU), Erlangen, Germany. He received his BE degree in 2013 at Department of Biomedical Engineering, College of Engineering, Peking University, Beijing, China.
He received the MS degree in 2016 from Pattern Recognition Lab, Department of Computer Science, FAU. Later he continued his PhD and postdoctoral research at Pattern Recognition Lab till April 2021. 
His current research interests include machine learning in medical imaging and radiation oncology.
\end{IEEEbiography}
\begin{IEEEbiography}[{\includegraphics[width=1in,height=1.25in,clip,keepaspectratio]{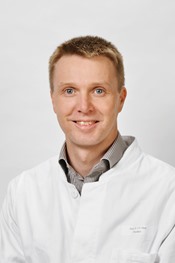}}]{Christoph Bert}
studied physics at Friedrich-Alexander Universit\"at Erlangen-N\"urnberg (FAU) and at Imperial College in London. He received the PhD from Technische Universit\"at (TU) Darmstadt in 2006. Since 2012, Dr. Bert is Professor of Medical Radiation Physics at FAU Erlangen-N\"urnberg, and head of medical physics at the Department of Radiation Oncology of the University Hospital Erlangen. His research focus aims at improving therapy techniques and their quality assurance. Current focus topics are the management of organ motion by tracking, surface guided radiation therapy (SGRT), MRI in radiation oncology, and error detection in interstitial brachytherapy.
\end{IEEEbiography}
\begin{IEEEbiography}[{\includegraphics[width=1in,height=1.25in,clip,keepaspectratio]{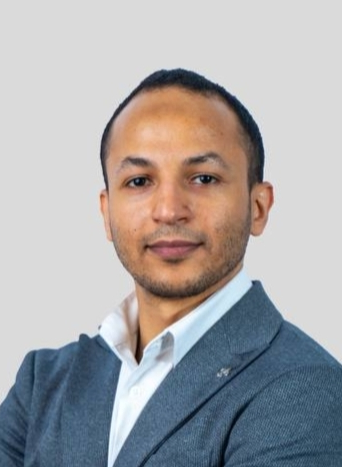}}]{Ahmed Gomaa}
is a Ph.D. candidate at Department of Radiation Oncology, University Hospital Erlangen, Friedrich-Alexander Universit\"at Erlangen-N\"urnberg (FAU), Germany. He received his BSc degree in 2017 at Department of Electrical Engineering at King Fahd University of Petroleum and Minerals, KSA. He received his MSc in Communication and Multimedia Engineering in 2022 at FAU. His research focuses on the application of deep learning in brain medical images.
\end{IEEEbiography}
\begin{IEEEbiography}[{\includegraphics[width=1in,height=1.25in,clip,keepaspectratio]{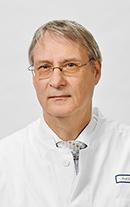}}]{Rainer Fietkau} is professor and chair of Department of Radiation Oncology, University Hospital Erlangen, Friedrich-Alexander Universit\"at Erlangen-N\"urnberg (FAU), Erlangen, Germany since 2008. He received the PhD degree in the Institute of Medical Radiation Science, University of W\"urzburg in 1985. In 1989, he became a senior physician at Department of Radiation Oncology, University Hospital Erlangen, FAU, where he finished his habilitation in 1992. In 1993, he was head of the Clinic and Polyclinic for Radiation Therapy, University G\"ottingen. From 1996 to 1997, he was professor in Department of Radiation Oncology, University Hospital Erlangen, FAU. From 1997 to 2008, he became professor at Department of Radiation Oncology, Radiation Therapy and Radiation Biology in University Rostock. Since 2014, he is speaker of the Research Group for Clinical application of hyperthermia in oncology. Since 2019, he has become president of the German Society for Radiation Oncology (DEGRO). His major scientific research interests lie in chemoradiation therapy, combination of radiotherapy with immunotherapy, and hyperthermia.
\end{IEEEbiography}
\begin{IEEEbiography}[{\includegraphics[width=1in,height=1.25in,clip,keepaspectratio]{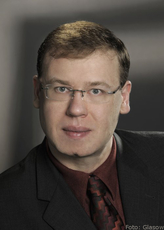}}]{Andreas Maier}
studied Computer Science, graduated in 2005, and received his PhD in 2009. From 2005 to 2009 he was working at the Pattern Recognition Lab at the Computer Science Department, Friedrich-Alexander Universit\"at Erlangen-N\"urnberg (FAU). His major research subject was medical signal processing in speech data. 
From 2009 to 2010, he started working on flat-panel C-arm CT as post-doctoral fellow at the Radiological Sciences Laboratory, Department of Radiology, Stanford University. From 2011 to 2012 he joined Siemens Healthcare as innovation project manager and was responsible for reconstruction topics in the Angiography and X-ray business unit.
In 2012, he returned to FAU as head of the Medical Reconstruction Group at the Pattern Recognition lab. In 2015 he became professor and head of the Pattern Recognition Lab. Since 2016, he is member of the steering committee of the European Time Machine Consortium. In 2018, he was awarded an ERC Synergy Grant “4D nanoscope”.  Current research interests focuses on medical imaging, image and audio processing, digital humanities, and interpretable machine learning and the use of known operators.
\end{IEEEbiography}
\begin{IEEEbiography}[{\includegraphics[width=1in,height=1.25in,clip,keepaspectratio]{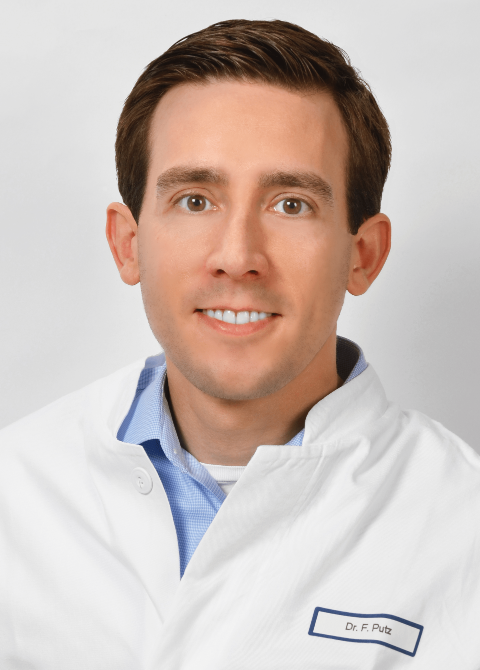}}]{Florian Putz} is a senior physician and working group leader at the Department of Radiation Oncology of the University Hospital Erlangen. He graduated in 2012 and received his MD from Friedrich-Alexander University Erlangen-N\"urnberg (FAU) in 2015. Florian Putz has initiated and is conducting multiple publicly and industrially-funded research projects including clinical trials, aiming to develop and evaluate deep learning applications for clinical requirements. His current research focus is on machine learning in medical imaging, imaging in radiation oncology and understanding the interplay between deep learning applications and human experts.
\end{IEEEbiography}

\vfill

\end{document}